\documentclass[twoside,11pt,xcolor={table}]{article}

 \usepackage[abbrvbib, preprint]{jmlr2e}

\usepackage[utf8]{inputenc} 
\usepackage[T1]{fontenc}    
\usepackage{hyperref}       
\usepackage{url}            
\usepackage{booktabs}       
\usepackage{amsfonts}       
\usepackage{nicefrac}       
\usepackage{microtype}      
\usepackage{times}
\usepackage{epsfig}
\usepackage{graphicx}
\usepackage{amsmath}
\usepackage{amssymb}
\usepackage{tc}
\usepackage{pdfrender}
\newcommand{\boldcheckmark}{%
  \textpdfrender{
    TextRenderingMode=FillStroke,
    LineWidth=.5pt, 
  }{\checkmark}%
}
\usepackage{hyperref}
\usepackage{url}
\hypersetup{colorlinks}
\usepackage{mathtools}
\usepackage{algorithm}
\usepackage{enumerate}
\usepackage[noend]{algpseudocode}
\usepackage{makecell}
\usepackage{comment}
\usepackage{subcaption}
\usepackage{enumitem}
\usepackage{arydshln}
\usepackage{bm}
\usepackage{siunitx,tabularx,ragged2e,booktabs}
\usepackage{wrapfig}
\usepackage{float}
\usepackage{booktabs}
\usepackage{pifont}
\usepackage{soul}
\usepackage{epstopdf}
\usepackage{fancyvrb}
\usepackage{multirow}
\usepackage{tabularx}
\usepackage{adjustbox}
\usepackage{cleveref}
\usepackage[dvipsnames,table]{xcolor}
\usepackage[english]{babel}

\usepackage{tabularray}
\UseTblrLibrary{varwidth, siunitx}

\usepackage{framed}
\usepackage{tikz}
\usetikzlibrary{arrows}
\usepackage{setspace}
\tikzset{
  treenode/.style = {align=center, inner sep=0pt, text centered,
    font=\sffamily},
  arn_n/.style = {treenode, circle, white, font=\sffamily\bfseries, draw=black,
    fill=black, text width=1.5em},
  arn_r/.style = {treenode, circle, red, draw=red,
    text width=1.5em, very thick},
  arn_x/.style = {treenode, rectangle, draw=black,
    minimum width=0.5em, minimum height=0.5em}
}
\fvset{frame=single,framesep=1mm,fontfamily=courier,fontsize=\scriptsize,numbers=left,framerule=.3mm,numbersep=1mm,commandchars=\\\{\}}

\usepackage[outline]{contour}
\contourlength{0.25pt}
\contournumber{20}

\newcommand{\ie}{\textit{i.e.}}
\newcommand{\eg}{\textit{e.g.}}

\newcommand{\dhspg}{\text{DHSPG{}}}
\newcommand{\hhspg}{\text{H2SPG{}}}
\newcommand{\otovone}{OTOv1{}}
\newcommand{\otovtwo}{OTOv2{}}
\newcommand{\algacro}{OTOv3{}}
\newcommand{\xcheckmark}{\checkmark\kern-1.1ex\raisebox{.7ex}{\rotatebox[origin=c]{125}{--}}}
\newcommand{\xmark}{\ding{55}}%

\newcommand{\cifar}{\text{CIFAR10}}
\newcommand{\cifarhundred}{\text{CIFAR100}}
\newcommand{\fashionmnist}{\text{Fashion-MNIST}}
\newcommand{\stl}{\text{STL-10}}
\newcommand{\svnh}{\text{SVNH}}

\newcommand{\resnetfifty}{\text{ResNet50}}
\newcommand{\demonet}{\text{DemoNet}}

\newcommand{\densenetonetwoone}{\text{DenseNet121}}
\newcommand{\stackedunets}{\text{StackedUnets}}
\newcommand{\superresnet}{\text{SuperResNet}}
\newcommand{\darts}{\text{DARTS}}
\newcommand{\vgg}{\text{VGG16}}
\newcommand{\vggbn}{\text{VGG16-BN}}

\newcommand{\bert}{\text{Bert}}
\newcommand{\squad}{\text{SQuAD}}
\newcommand{\regnet}{\text{RegNet}}

\newcommand{\imagenet}{\text{ImageNet}}

\usepackage{lastpage}
\jmlrheading{23}{2023}{1-\pageref{LastPage}}{12/23; Revised --}{--}{21-0000}{Tianyi Chen}

\ShortHeadings{OTOv3: Automatic Prune and Erase Operators in DNN}{Chen, Ding, et al.}
\firstpageno{1}

\begin{document}

\title{OTOv3: Automatic Architecture-Agnostic Neural Network Training and Compression from Structured Pruning to Erasing Operators}

\author{\noindent
\name Tianyi Chen \email tiachen@microsoft.com \\
\addr Microsoft\\
Redmond, WA 98052, USA
\AND
\name Tianyu Ding \email tianyuding@microsoft.com \\
\addr Microsoft\\
Redmond, WA 98052, USA
\AND
\name Zhihui Zhu \email zhu.3440@osu.edu \\
\addr  The Ohio State of University\\
Columbus, OH 43210, USA
\AND
\name Zeyu Chen \email zeyu.chen@microsoft.com \\
\addr Microsoft\\
Redmond, WA 98052, USA
\AND
\name HsiangTao Wu \email musclewu@microsoft.com \\
\addr Microsoft\\
Redmond, WA 98052, USA
\AND
\name Ilya Zharkov \email zharkov@microsoft.com \\
\addr Microsoft\\
Redmond, WA 98052, USA
\AND
\name Luming Liang \email lulian@microsoft.com \\
\addr Microsoft\\
Redmond, WA 98052, USA
}

\editor{My editor}

\maketitle

\begin{abstract}
Compressing a predefined deep neural network (DNN) into a compact sub-network with competitive performance is crucial in the efficient machine learning realm. This topic spans various techniques, from structured pruning to neural architecture search, encompassing both \textit{pruning} and  \textit{erasing} operators perspectives. Despite advancements, existing methods suffers from complex, multi-stage processes that demand substantial engineering and domain knowledge, limiting their broader applications. We introduce the third-generation Only-Train-Once (\algacro), which \textit{first automatically}  trains and compresses a general DNN through \textit{pruning} and \textit{erasing} operations, creating a compact and competitive sub-network without the need of fine-tuning. \algacro{} simplifies and automates the training and compression process, minimizes the engineering efforts required from users. It offers key technological advancements: \textit{(i)} automatic search space construction for general DNNs based on dependency graph analysis; \textit{(ii)} Dual Half-Space Projected Gradient (\dhspg{}) and its enhanced version with hierarchical search (\hhspg{}) to reliably solve (hierarchical) structured sparsity problems and ensure sub-network validity; and \textit{(iii)} automated sub-network construction using solutions from \dhspg/\hhspg{}  and dependency graphs. Our empirical results demonstrate \algacro's effectiveness across various benchmarks in structured pruning and neural architecture search.
\algacro{} produces sub-networks that match or exceed the 
state-of-the-arts.
The source code will be available at~\url{https://github.com/tianyic/only_train_once}.
\end{abstract}

\begin{figure}[t]
    \centering    
    \begin{minipage}{.58\textwidth}
    \centering
     \includegraphics[width=\linewidth]{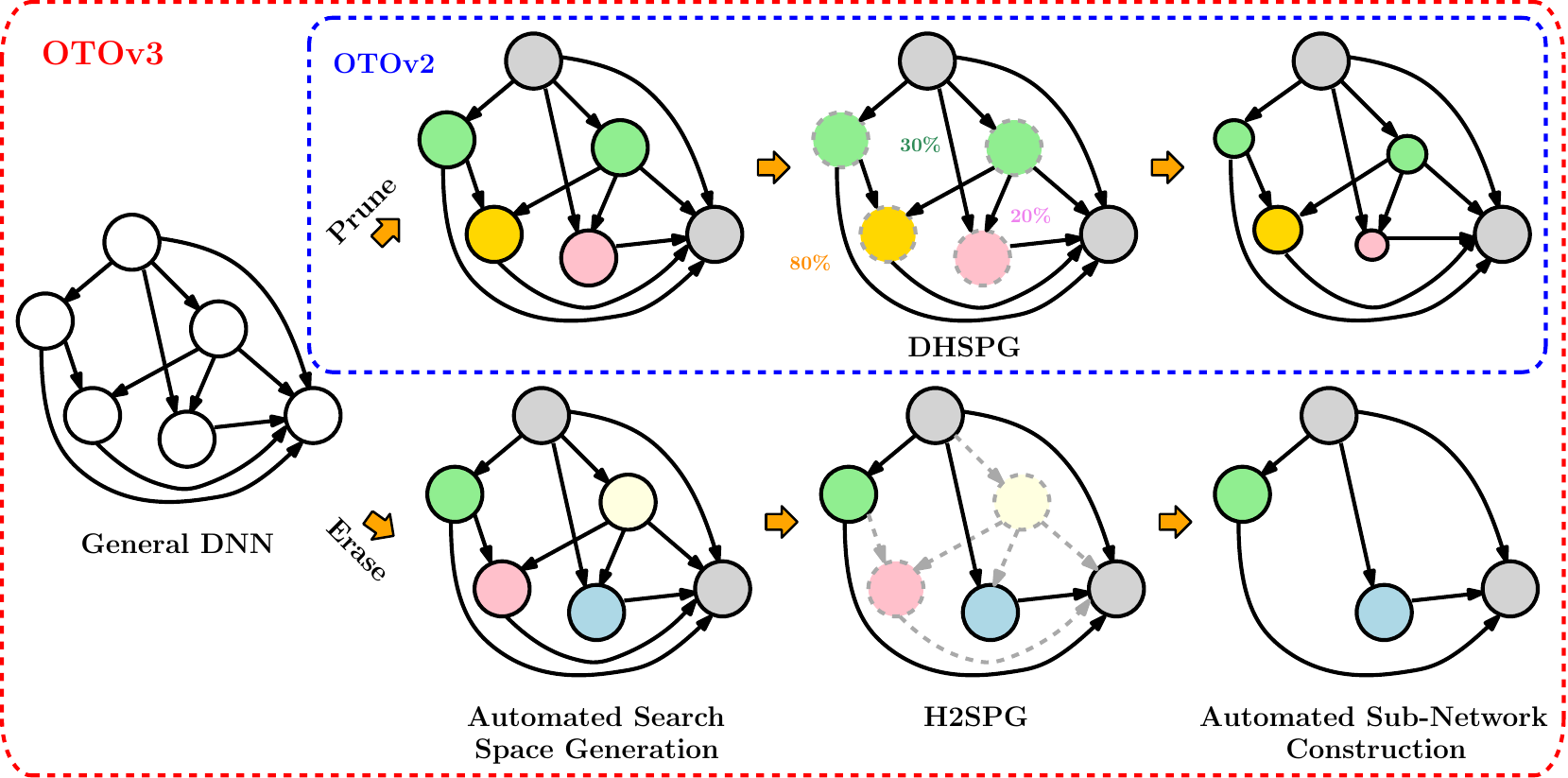}
    \end{minipage}
    \hspace{1.5mm}
\begin{minipage}{.39\textwidth}
\begin{Verbatim}[label={\textbf{OTOv3 Library Usage}}]
\textcolor{magenta}{from} only\_train\_once \textcolor{magenta}{import} OTO
\textcolor{Green}{\# Target general DNN}
oto = OTO(\textcolor{blue}{dnn}, \textcolor{blue}{compress_mode})
\textcolor{Green}{\# Structurally prune operators}
optimizer = oto.dhspg()
\textcolor{Green}{\# Erase operators}
optimizer = oto.h2spg()
\textcolor{Green}{\# Train as normal}
optimizer.step()
\textcolor{Green}{\# Construct sub-network}
oto.construct_subnet()
\end{Verbatim}
\end{minipage}
    \caption{
    Overview of \algacro{}. \algacro{} is the \textit{first} to support two modes of \textit{architecture-agnostic} and \textit{automated} training and compression, \ie, \textit{pruning} and \textit{erase} modes, to deliver high-performing compact sub-networks starting from a predefined DNN. \textit{(i)} Pruning mode slims the operators via eliminating the inherent redundant structures, yet keeps the presence of operators and connections. \textit{(ii)} Erasing mode removes the redundant operators entirely to dramatically simplify the DNN architecture. The library is user-friendly that requires \textit{minimal} engineering efforts from the end-users.} 
    \label{fig:overview}
\end{figure}

\section{Introduction}

Large-scale Deep Neural Networks (DNNs) have proven to be highly effective across various applications~\citep{he2016deep}. However, deploying these substantial networks in environments with limited resources presents significant challenges. As a result, both academic and industry increasingly focus on compressing DNNs into more compact versions with minimal performance loss. Despite a decade of research in the compression field, the complete  resolution remains an ongoing endeavor.

From the view of computational graph, the compression of DNNs could be largely considered as either \textit{\underline{pruning}} or \textit{\underline{erasing}} the existing operators to construct high-performing compact sub-networks. The former one is mainly studied via the structured pruning realm, which preserves the pre-defined architecture, yet slims each operator~\citep{chen2021oto,chen2023otov2,fang2023depgraph}. The latter one is mainly studied via the gradient-based and several zero-shot Neural Architecture Search (NAS) methods that identify less important operators from a super-network that covers all candidate operators and connections~\citep{liu2018darts,yang2020ista,xu2019pc,chen2021progressive}. 

Despite the advancements in both structured pruning and NAS methods, their usage is still limited due to certain inconvenience. In particular, these methods predominantly rely on manually determining the search space (see formal definitions in Section~\ref{sec.preliminary}) for a pre-specified DNN beforehand, and require the manual establishment of the whole searching / pruning pipeline. The whole process necessitates significant domain-knowledge and engineering efforts, thereby being inconvenient and time-consuming for users. Therefore, it is natural to ask whether we could reach an

\noindent
\textit{\textbf{Objective.} Given a \underline{general DNN} that \underline{covers all candidate operators and connections}, automatically generate its search space, train it once, and construct a high-performing compact sub-network.}

Achieving the objective is severely challenging in terms of both engineering developments and algorithmic designs, consequently rarely achieved by the existing structured pruning and NAS works to our knowledge\footnote{For automatic pruning, OTOv2~\citep{chen2023otov2} and DepGraph~\citep{fang2023depgraph} are arguably the first two works which appeared almost at the same time to achieve architecture-agnostic structured pruning.}. We now build the third generation of Only-Train-Once~(\algacro{}) that first reaches the objective from both pruning and erasing manners. Given a DNN that covers all operation and connection candidates, \algacro{} automatically generates search spaces for either pruning or erasing operator mode, trains and identifies redundant minimally removal structures, then builds a sub-network that achieves both high performance and compactness, as Figure~\ref{fig:overview}. The whole procedure is automatically proceeded, dramatically reduce human efforts, and fits general DNNs and applications. 

\begin{wraptable}{r}{0.4\textwidth}
	\vspace{-2mm}
	\resizebox{0.4\textwidth}{!}{
		\begin{tabular}{l|c|c|c}
			\Xhline{3\arrayrulewidth}
			&  \textbf{OTOv3} & \textbf{Other Pruning} & \textbf{Other NAS}$^\star$  \\
			\hline
			\textbf{General DNNs} & \textcolor{red}{\boldcheckmark} & \xcheckmark & \xcheckmark \\
			\textbf{Autonomy} & \textcolor{red}{\boldcheckmark} & \hspace{1.2mm}\xmark$^\dagger$ & \hspace{1.2mm}\xmark$^\ddagger$ \\
			\textbf{Prune Operations} & \hspace{1.2mm}\textcolor{red}{\boldcheckmark} & \textcolor{red}{\boldcheckmark} & \xmark \\
			\textbf{Erase Operations} & \textcolor{red}{\boldcheckmark} & \xmark & \textcolor{red}{\boldcheckmark} \\
			\hline
			\Xhline{3\arrayrulewidth}
			\multicolumn{4}{l}{$^\star$ Refer to gradient-based and a subset of zero-shot NAS.}\\
			\multicolumn{4}{l}{$^\dagger$ Except Torch-Pruning~\citep{fang2023depgraph}.}\\
			\multicolumn{4}{l}{$^\ddagger$ Require manually construct search space.}
	\end{tabular}}
	\vspace{-4mm}
\end{wraptable}
Technologically, this paper serves as a \textit{non-trivial} and \textit{significant} extension of \otovtwo{}~\citep{chen2023otov2} (automated structured pruning) to present a unified framework, enables architecture-agnostic DNN training and compression from both pruning and erasing operator perspectives. Compared with~\otovtwo{}, \algacro{} explores an \underline{\textit{entirely new application domain}} to automatically erase redundant operators, which presence are preserved in structured pruning, along with significant novel technical and practical contributions from both \underline{\textit{algorithmic designs}} and \underline{\textit{infrastructure developments}}. \algacro{} also bridges to the gradient-based NAS field and resolves one of their main pain points of handcrafting search space beforehand upon a pre-defined DNN. 
In the remaining of the paper, we will summarize the main context from \otovtwo{} with more insights, yet concentrate and highlight more on the novel takeaways in the erasing operator feature. Our main contributions can be summarized as follows.

\begin{itemize}[leftmargin=*]
\item \textbf{Infrastructure for Automated Architecture-Agnostic DNN Training and Compression.} We propose \algacro{} that is the first to \textit{\underline{automatically}} train and compress a general DNN to deliver a compact sub-network by \underline{\textit{pruning}} and \underline{\textit{erasing}} redundant operations in the one-shot manner. As the previous OTO versions, \algacro{} trains the DNNs only once  without the need of pre-training and fine-tuning, and is pluggable into various deep learning applications. 

\item \textbf{Automated Search Space Generation via Dependency Graph Analysis.}  We propose distinct novel graph algorithms to automatically exploit the architecture and establish pruning and erasing dependency graphs for separate compression purpose given a predefined DNN. The dependency graphs analyze the hierarchy and dependency across different operators to form search spaces for different compression modes. The search spaces consist of the pruning or erasing minimally removal structures that could be removed without interrupting the functionality of the sub-networks. 

\item \textbf{Dual Half-Space Projected Gradient (\dhspg{}) for Pruning Mode.} We propose a novel sparse optimizer \dhspg{}, to train and structurally prune a general DNN. \dhspg{} formulates a constrained structured sparse optimization problem and solves it by modular redundant identification and a hybrid training schema. Compared with HSPG in OTOv1 and other sparse optimizers, \dhspg{} outperforms them in terms of more reliably sparsity control and better generalization.

\item \textbf{Hierarchical Half-Space Projected Gradient (\hhspg{}) for Erasing Mode.} We further propose a novel \hhspg{} for erasing operators. \hhspg{} is perhaps the first optimizer, that solves a hierarchical structured sparsity problem for general DNN applications. Compared to \dhspg{} and other sparse optimizers, \hhspg{} conducts a dedicated hierarchical search phase over the generated erasing search space to ensures the validness of the sub-network after erasing redundant operators.

\item \textbf{Automated Sub-Network Construction.} We propose novel graph algorithms to automatically construct sub-networks upon the solution of \dhspg{} or \hhspg{} for pruning and erasing operator perspectives. The resulting sub-network returns the exact same outputs as the full networks thereby being no need of further fine-tuning. 

\end{itemize}

\section{Related Works}\label{sec.related_work}

\paragraph{Structured Pruning.} To compress DNN architectures, structured pruning shrinks the operators while preserves their presence via identifying and removing redundant inherent structures~\citep{gale2019state, han2015deep,lin2019toward,wen2016learning}. The general procedure ~\citep{li2023model} can be largely summarized as: (\textit{i}) train a full model; (\textit{ii}) identify and remove the redundant structures to construct a slimmer DNN based on various criteria, including sparsity~\citep{lin2019toward,wen2016learning,li2020group,zhuang2020neuron,chen2017reduced,chen2018farsa,chen2021orthant,chen2020neural,gao2020highly,zhuang2020neuron,meng2020pruning,yang2019deephoyer,frantar2023sparsegpt,idelbayev2022exploring}, Bayesian pruning~\citep{zhou2019accelerate,van2020bayesian}, ranking importance~\citep{li2020eagleeye,hu2016network,li2019exploiting,zhang2018systematic}, grouped kernel search~\cite{zhong2023one}, spectral graph analysis~\citep{laenen2023one}, reinforcement learning~\citep{he2018amc,chen2019storage},~lottery ticket~\citep{frankle2018lottery,frankle2019stabilizing,renda2020comparing}, etc.; (\textit{iii}) retrain the pruned model to regain the accuracy regression during pruning if needed with(out) knowledge distillation~\citep{ham2023cosine}. These methods have to conduct a complicated and time-consuming procedure to trains the DNN multiple times and requires a good deal of
domain knowledge to manually proceed every individual step. OTOv1~\citep{chen2021oto} is then proposed to avoid fine-tuning and end-to-end train and compress the DNN once. However, these methods requires numerous handcrafting efforts on discovering the removal structures and constructing slimmer model for specific DNNs in advance, thereby is not convenient. Recent methods such as OTOv2~\citep{chen2023otov2} and DepGraph~\citep{fang2023depgraph} made progress in automating the structure pruning process for general DNNs. OTOv2 is a one-shot training and pruning framework that does not need pre-training or fine-tuning. DepGraph focuses on pruning, hence requires integrating with a multi-stage training pipeline. In this work, we propose the third-generation version of OTO that significantly enhances the existing pruning mode of OTOv2 from the engineering perspective and extends it to a new application domain to automatically erase redundant operators entirely.

\paragraph{Neural Architecture Search (NAS) for Erasing Operators.} Erasing redundant operators in a pre-defined DNN to search an optimal sub-network has been studied as a pivotal topic in the NAS realm. There exists gradient-based methods~\citep{liu2018darts,chen2019progressive,xu2019pc,yang2020ista,hosseini2022saliency} and zero-shot methods~\citep{chen2021neural,li2023zico} that start with a network covering all possible connection and operation candidates to find important operators to form a sub-network. However, similarly to the structured pruning realm, these methods require a significant amount of \textit{handcraftness} from users in advance to \textit{manually} establish the search space, introduce additional architecture variables, and build the multi-level training pipeline. The sub-network construction is also network-specific and not flexible. All requirements necessitate remarkable domain-knowledge and expertise, making it difficult to extend to broader scenarios.

\paragraph{Automated Search Space Generation.} One main pain-point of the existing structured pruning and NAS methods is the need of \textbf{manually} establishing the search space. The definition of search space is varying upon different scenarios. In our scenario, we aim to automatically discovering a high-performing compact sub-network given a heavier general DNN. The starting DNN is assumed to cover all operation and connection candidates, and the resulting sub-network serves as its sub-computational-graph. Therefore, the search space of our scenario is defined as a set of minimally removal structures of the given DNN in distinct compression modes (see Definition~\ref{def:search_space}).  

There exists \textit{orthogonal} search-space definitions, along with works in automation. In the context of \citep{cai2019once,munoz2022automated,radosavovic2020designing,calhas2022automatic}, the presence of operators in DNNs is preserved, yet their inherent hyperparameters, such as stride and depth for convolution, are searchable. Consequently, the inherent hyperparameters of the operators constitute their search spaces. \cite{zhou2021autospace} defines the search space as the network that encompasses all candidate operations and investigates to automatically generate high-quality super-networks that include optimal sub-networks. \algacro{} studies the automation for two distinct and important search spaces to prune and erase operators. It stays complementary to other methods and softwares and could operate jointly to form the landscape of automated search-space generation and automated machine learning~\citep{pmlr-v123-liu20a,xu2021automl,pedregosa2011scikit,kolter2021xrl}.
\section{Preliminaries}\label{sec.preliminary}

We review preliminary concepts used throughout the paper.  
To compress a DNN, we need to identify or search a set of redundant structures, removing them to construct a compact high-performing sub-network. Due to the complicated connectivity of DNNs, removing an arbitrary structure may result in an invalid DNN. Consequently, the first concept is so-called \textit{removal structure}. 
\begin{definition}[Removal Structure]\label{def:removal_structure}
Given a deep neural network $\mathcal{M}$, a structure is said removal structure if and only if the neural network without it is still functioning normally.
\end{definition}
One removal structure may contain multiple individually instances. 
Searching redundancy over a large removal structure may result in sub-optimal sub-network.  To enlarge search space, we further introduce the \textit{minimally removal structure} that describes the minimal units of one removal structure.  
\begin{definition}[Minimally Removal Structure]\label{def:minimal_removal_structure}
Given a deep neural network $\mathcal{M}$, a minimally removal structure is a removal structure that can not decompose into multiple removal structures. 
\end{definition} 
Now, we formally define the search space of compressing DNN via pruning and erasing operators. 
\begin{definition}[Search Space for DNN Compression]\label{def:search_space}
Given a DNN $\mathcal{M}$, the search space for training and compression $\mathcal{M}$ via pruning and erasing operators is the set of minimally removal structures. 
\end{definition}
The definition of search space is varying upon distinct scenarios (see more in Section~\ref{sec.related_work}).  In our compression context, the search space is defined as the set of minimally removal structures that can be omitted from the given DNN while ensuring that the remaining network continues to function normally. Furthermore, to train \textit{only once} without the need of fine-tuning after compression, we introduce the zero-invariant group (ZIG) in \otovone~\citep{chen2021oto}. 
\begin{definition}[Zero Invariant Group]\label{def:zero_invariant_group}
The trainable variables of one minimally removal structure form a zero-invariant group if and only if these trainable variables being zero results in the minimally removal structure always returning output tensors as zero given arbitrary inputs. 
\end{definition}
ZIG describes a class of minimally removal structures such that their variables parameterized as zero ensure the output tensors as zero, thereby contributing none to the DNN outputs. During DNN compression, the redundant ZIGs are eventually projected onto zero, which structures could be further removed without affecting the model output, thereby avoiding the need of further fine-tuning. 
\paragraph{Remark.} Upon different compression purpose, the form of minimally removal structure is \textit{varying} and requires distinct dedicately designed graph algorithms to automatically discover. For clarification, we introduce \textbf{\textit{pruning minimally removal structures}} and \textbf{\textit{erasing minimally removal structures}} for the pruning and erasing compression modes, respectively. Similarly, we have \textbf{\textit{pruning search space}} and \textbf{\textit{erasing search space}} along with \textbf{\textit{pruning zero-invariant groups (PZIGs)}} and \textbf{\textit{erasing zero-invariant groups (EZIGs)}} to distinguish separate compression modes.

\section{Overview of \algacro{}}\label{sec.overview}

\begin{algorithm}[t]
	\caption{Outline of \algacro{}.}
	\label{alg:main.outline}
	\begin{algorithmic}[1]
		\State \textbf{Input:} A DNN $\mathcal{M}$ to be trained and 
        compressed (no need to be pretrained), a compression mode $\in \{\text{prune}, \text{erase}\}$, and a target sparsity level $K$.
		\State \textbf{Automated Search Space Generation.} Create dependency graphs for $\mathcal{M}$, generate a search space upon the compression mode, and partition the trainable variables into a set of groups $\mathcal{G}$. 
		\State \textbf{Train $\mathcal{M}$ by \dhspg{} or \hhspg{}.} Select the sparse optimizer upon the predefined compression mode. Then seek a high-performing solution with desired (hierarchical) group sparsity $K$. 
		\State \textbf{Automated Sub-Network Construction.} Construct a sub-network $\mathcal{M}^*$ upon solution of \dhspg{} or \hhspg{} and the selected compression mode.
		\State \textbf{Output:} Constructed sub-network $\mathcal{M}^*$. (Post fine-tuning is optional).
 \end{algorithmic}
\end{algorithm}

Given a deep neural network $\mathcal{M}$, \algacro{} establishes a unified paradigm to conduct training and compression via both structured pruning and erasing operators as diagrammed in Figure~\ref{fig:overview}. As outlined in Algorithm~\ref{alg:main.outline}, the end-users firstly need to select a compression mode via either pruning or erasing. Upon the selected mode, \algacro{} then automatically analyzes the relationships among the operators via distinct dependency graph algorithms to establish corresponding search space~(Section~\ref{sec.search_space_pruning} or~\ref{sec.search_space_erasing}). The search spaces consist of minimally removal structures (Definition~\ref{def:search_space}) for different purposes. Their trainable variables are then partitioned into different sets of groups (including zero-invariant groups (ZIGs) and the complementary) $\mathcal{G}_\text{prune}$ and $\mathcal{G}_\text{erase}$ for pruning or erasing mode, respectively. 

To identify redundant minimally removal structures and train the important ones for high-performance, separately distinct structured sparsity optimization problems are formulated for pruning and erasing mode over $\mathcal{G}_\text{prune}$ and $\mathcal{G}_\text{erase}$, respectively. For pruning mode, the optimization problem is a \textit{disjoint} group sparse problem and solved via a Dual Half-Space Projected Gradient (\dhspg{}) method to yield a solution $\bm{x}^*_{{\dhspg}}$ with competitive performance as well as high group sparsity in the view of $\mathcal{G}_\text{prune}$ (Section~\ref{sec.dhspg}). For erasing mode, the problem becomes a more challenging \textit{hierarchical} group sparse problem, wherein the sparsity must be yielded obeying the hierarchy to ensure the validity of the remaining network. We propose a novel Hierarchical Half-Space Projected Gradient (\hhspg{}) to effectively solve it and return a solution $\bm{x}^*_{{\hhspg}}$ with both high-performance and desired hierarchical sparsity (Section~\ref{sec.hhspg}). 
In the end, sub-networks $\mathcal{M}^*_\text{prune}$ and $\mathcal{M}^*_\text{erase}$ are automatically constructed by removing corresponding redundant structures upon the selected mode and the solutions $\bm{x}^*_{{\dhspg}}$ and $\bm{x}^*_{{\hhspg}}$ from \dhspg{} and \hhspg{} (Section~\ref{sec.sub_network_pruning} and~\ref{sec.sub_network_erasing}). Post fine-tuning is optional and typically no needed especially if all variable groups are all zero-invariant groups. The whole procedure is proceeded automatically and easily employed onto various DNN applications, and consumes almost minimal engineering efforts from the users. 

\begin{figure}[t]
    \centering
    \includegraphics[width=\linewidth]{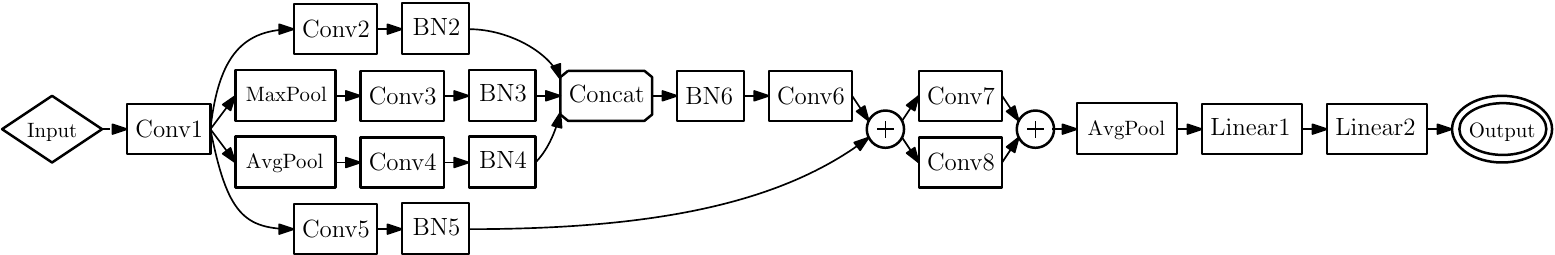}
	\caption{A demo DNN (\demonet{}) to be trained and compressed.}
    \label{fig:demonet_original}
\end{figure}

\section{\algacro{} for Structured Pruning}\label{sec.pruning_mode}

The first application domain of \algacro{} is to automatically train a general DNN only once and simultaneously structurally prune the operators to deliver a high-performing compact sub-network. The main context has been covered in \otovtwo{}~\citep{chen2023otov2}. In this section, we reorganize the method presentation with {\textit{deeper insights}} and {\textit{more elaborations}}. 

\subsection{Automatic Pruning Search Space Generation for Pruning Mode}\label{sec.search_space_pruning}

\begin{figure}[t]
    \centering
    \begin{subfigure}{\linewidth}
    	\centering
    	\includegraphics[width=\linewidth]{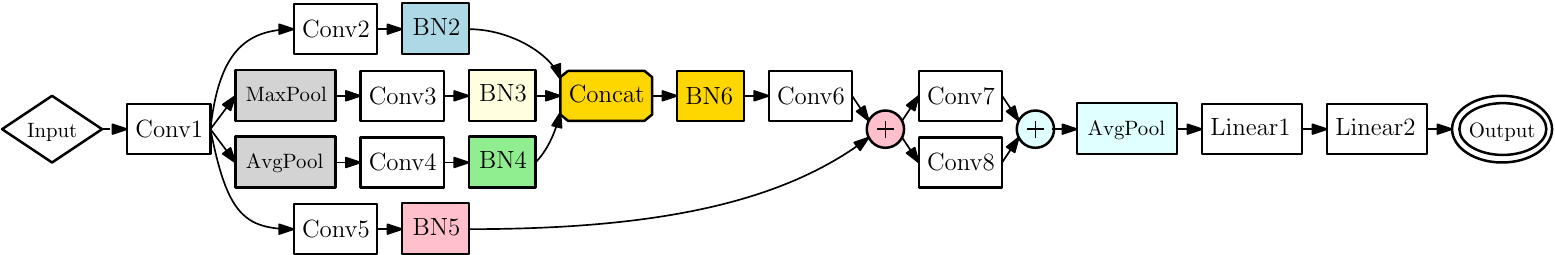}
    	\caption{During constructing pruning dependency graph.}\label{fig:demonet_dependency_graph_skeleton_pruning}
    	\vspace{1mm}
    	\label{fig:1a}		
    \end{subfigure}
    \begin{subfigure}{\linewidth}
    	\centering
    	\includegraphics[width=\linewidth]{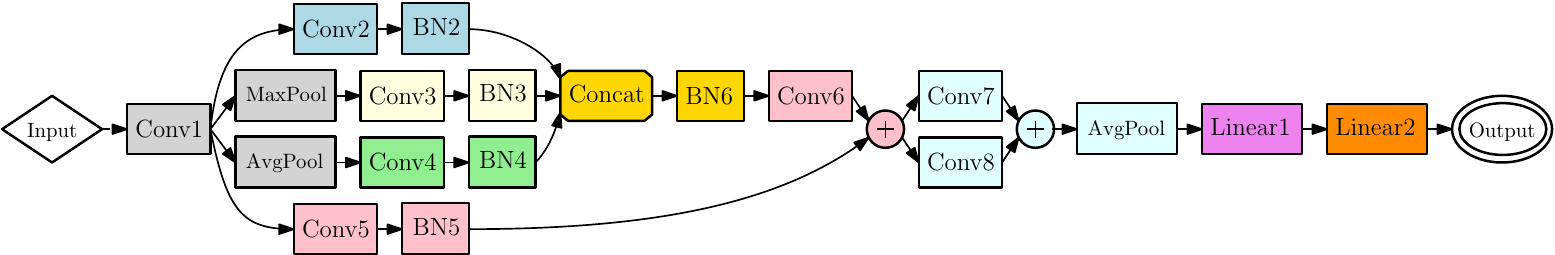}
    	\caption{Pruning dependency graph.}\label{fig:demonet_dependency_graph_pruning}
    	\vspace{1mm}
    \end{subfigure}
    \begin{subfigure}{\linewidth}
    	\centering
    	\includegraphics[width=\linewidth]{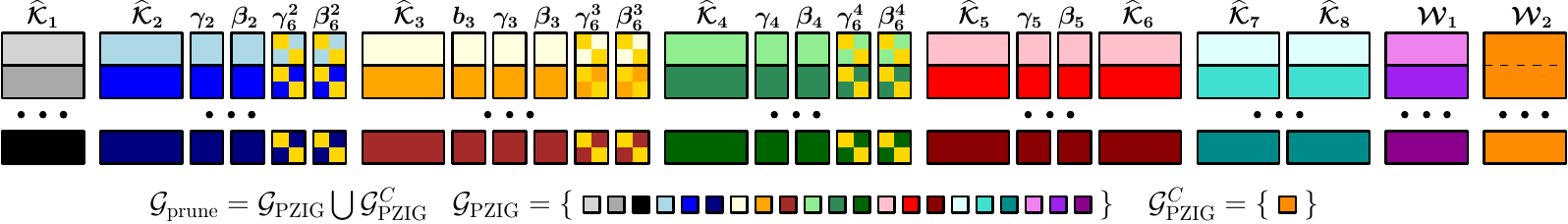}
    	\caption{Pruning zero-invariant groups (PZIGs) and the complementary.}\label{fig:demonet_zigs_pruning}
    \vspace{-3mm}
    \end{subfigure}
    \caption{Automated PZIG partition. $\widehat{\bm{\mathcal{K}}}_i$ and $\bm{b}_i$ are the flatten filter matrix and bias vector of \texttt{Conv$i$}, where the $j$th row of $\widehat{\bm{\mathcal{K}}}_i$ represents the $j$th 3D filter. $\bm{\gamma}_i$ and $\bm{\beta}_i$ are the  weighting and bias vectors of \texttt{BN$i$}. $\bm{\mathcal{W}}_i$ is the weighting matrix for \texttt{Linear$i$}.
    PZIGs are present in Figure~\ref{fig:demonet_zigs_pruning}. Since the output tensors of \texttt{Conv5} and \texttt{Conv6} are added together, both layers associated with \texttt{BN5} must remove the same number of filters from $\widehat{\bm{\mathcal{K}}}_5$ and $\widehat{\bm{\mathcal{K}}}_6$ and scalars from $\bm{\gamma}_5, \bm{\beta}_5$ to keep the addition valid. Since \texttt{BN6} normalizes the concatenated outputs along channel from \texttt{Conv2-BN2} to \texttt{Conv4-BN4}, the corresponding scalars in $\bm{\gamma}_6, \bm{\beta}_6$ need to be grouped into these node groups accordingly.
    }
    \label{fig:automatic_search_space_construction} 
    \vspace{-4mm}
\end{figure}

As described in Section~\ref{sec.overview}, the foremost step is to establish a search space for structured pruning that consists of pruning minimally removal structures. Unlike the existing works that manually find them out, we propose a graph algorithm to achieve automated construction, stated in Algorithm~\ref{alg:search_space_pruning}. We elaborate the algorithm over a shallow yet complex \demonet{} depicted in Figure~\ref{fig:demonet_original}.

\paragraph{Roles of Vertices / Operators.}  We begin by constructing the trace graph $(\mathcal{E}, \mathcal{V})$ for the target DNN. In this graph, each vertex in $\mathcal{V}$ represents a specific operator, while the edges in $\mathcal{E}$ illustrate the connections between these operators (line~\ref{line:vertices_edges_trace_graph} in Algorithm~\ref{alg:search_space_pruning}). The vertices are categorized into four types: {\textit{stem}}, {\textit{joint}}, {\textit{accessory}}, and {\textit{unknown}}. \textit{Stem} vertices, equipped with trainable parameters, are capable of transforming input tensors into different shapes. Examples include operators like \texttt{Conv} and \texttt{Linear}. \textit{Joint} vertices, such as \texttt{Add}, \texttt{Mul}, and \texttt{Concat}, combine multiple input tensors into a single output. \textit{Accessory} vertices manipulate a single input tensor to produce a single output and may also have trainable parameters, like \texttt{BatchNorm} and \texttt{ReLu}. Lastly, \textit{unknown} vertices perform unseen operations. Notably, stem vertices constitute the majority of the DNN's parameters. Joint vertices, on the other hand, create connections across different vertices, adding hierarchy and complexity to the DNN. To maintain the integrity of the joint vertices, minimal removal structures must be meticulously devised. Moreover, joint vertices are further categorized based on their dependency on input shape. Vertices requiring inputs of identical shapes, such as \texttt{Add}, are labeled as shape-dependent (SD), otherwise being shape-independent (SID) such as \texttt{Concat} along the channel dimension for multiple \texttt{Conv} layers as input.

\begin{algorithm}[h]
	\caption{Automated Pruning Search Space Construction.}
	\label{alg:search_space_pruning}
	\begin{algorithmic}[1]
		\State \textbf{Input:} A DNN $\mathcal{M}$ to be trained and compressed. 
		\State Construct the trace graph $(\mathcal{V}, \mathcal{E})$ of $\mathcal{M}$.\label{line:vertices_edges_trace_graph}
		\State Find node groups $\mathcal{N}$ over all accessory, shape-dependent joint and unknown vertices.\label{line:cc_non_stem_vertices}
		\State Grow $\mathcal{N}$ till incoming nodes are either stem or shape-independent joint vertices.
		\label{line:grow_cc}
		\State Merge node groups in $\mathcal{N}$ if any intersection.\label{line:merge_cc}
		\State Group pairwise parameters of stem vertices in the same node groups associated with parameters from affiliated accessory vertices if any as one PZIG into $\mathcal{G}_\text{PZIG}$.
        \State Group parameters in the remaining vertices into $\mathcal{G}_\text{PZIG}^C$.
		\State \textbf{Return} variable groups partitions $\mathcal{G}_\text{prune}\gets \mathcal{G}_\text{PZIG}\cup \mathcal{G}_\text{PZIG}^C$.
	\end{algorithmic}
\end{algorithm}

\paragraph{Pruning Dependency Graph.}  To identify the pruning minimal removal structures in the pruning mode of the target DNN, we begin by analyzing the dependencies between the vertices. Our first step is to connect accessory, shape-dependent (SD) joint, and unknown vertices that are adjacent to form a set of node groups $\mathcal{N}$ (line~\ref{line:cc_non_stem_vertices} of Algorithm~\ref{alg:search_space_pruning}). This step lays the foundation for identifying interdependent vertices when contemplating the removal of hidden structures. The rationale behind this step is threefold: \textit{(i)} the adjacent accessory vertices operate and are subject to the same ancestral stem vertices if any; \textit{(ii)} SD joint vertices force their ancestral stem vertices dependent on each other to yield tensors in the same shapes;  and \textit{(iii)} unknown vertices introduce uncertainty, making it essential to identify potentially affected vertices. Following this, we expand $\mathcal{N}$ to include all incoming vertices that are either stem or shape-independent (SID) joint vertices (lines~\ref{line:grow_cc}-\ref{line:merge_cc}). Remark here that the newly incorporated stem vertices are associated with the accessory vertices. For example, in Figure~\ref{fig:demonet_dependency_graph_skeleton_pruning}, \texttt{Conv2} is affiliated with \texttt{BN2}, and \texttt{Conv3} is affiliated with \texttt{BN3}. Furthermore, SID joint vertices play a significant role in establishing dependencies. They not only link their affiliated accessory vertices but also connect with incoming components. For instance, \texttt{Concat-BN6}'s dependency on \texttt{Conv2-BN2}, \texttt{Conv3-BN3}, and \texttt{Conv4-BN4} arises because \texttt{BN6} normalizes the concatenated tensors along the channel dimension. Finally, the pruning dependency graph is constructed as Figure~\ref{fig:demonet_dependency_graph_pruning}, wherein the vertices in the same node group (marked as the same color) imply the interdependency during structured pruning to ensure the validity of produced subnetwork.

\paragraph{Pruning Zero-Invariant Group Partition.} We finally partition the trainable variables upon the pruning minimally removal structures into pruning zero-invariant groups (PZIGs) based on the pruning dependency graph. The process begins by grouping together the pairwise trainable variables of all individual stem and accessory vertices within the same node group (marked by the same color in Figure~\ref{fig:demonet_zigs_pruning}). Some accessory vertices like \texttt{BN6} may rely on multiple groups due to their connections with SID joint vertices. Consequently, the trainable variables are divided and added separately into the corresponding groups, \eg, $(\bm{\gamma}_6^2, \bm{\beta}_6^2)-(\bm{\gamma}_6^4, \bm{\beta}_6^4)$. 
Furthermore, we exclude node groups adjacent to DNN output from forming PZIGs to preserve the output shapes, as exemplified by \texttt{Linear2}. For safety and to ensure the framework generality when applied to DNNs with custom operators, node groups containing unknown vertices are also excluded due to the uncertainty associated with these vertices. Finally, the pruning zero-invariant groups $\mathcal{G}_\text{PZIG}$ are unioned with the remaining unprunable variable groups $\mathcal{G}_\text{PZIG}^C$ to form the overall variable partition $\mathcal{G}_\text{prune}$ in the pruning mode. 

\subsection{Dual Half-Space Projected Gradient (DHSPG)}\label{sec.dhspg}

\paragraph{Target Problem.} Given the variable partition $\mathcal{G}_\text{prune}$ by Algorithm~\ref{alg:search_space_pruning}, the next is to jointly search which groups in $\mathcal{G}_\text{PZIG}$ are redundant to be removed and train the remaining groups to achieve high performance. To tackle it, we construct a structured sparsity optimization problem formulated as~\eqref{prob.main.prune}. Besides minimizing the objective function $f$, we introduce an additional sparsity constraint to yield $K$ of the PZIGs being zero, wherein the zero groups refer to the redundant structures, and the non-zero groups exhibit the prediction power to maintain competitive performance to the full model.
\vspace{-1mm}
\begin{equation}\label{prob.main.prune}
\minimize{\bm{x}\in \mathbb{R}^n}\ f(\bm{x}),\ \ \text{s.t.} \ \text{Cardinality}\{g\ |\  [\bm{x}]_g=\bm{0}, g\in\mathcal{G}_\text{PZIG} \}=K. 
\vspace{-1mm}
\end{equation}
Problem~\eqref{prob.main.prune} is solved via a novel DHSPG (outlined as Algorithm~\ref{alg:main.dhspg}). Compared with HSPG~\citep{chen2020half,chen2021oto}, DHSPG employs saliency-driven redundant identification and a \underline{\textit{hybrid training}} paradigm to more reliably control the sparsity and achieve better generalization performance. 
\vspace{1mm}

\begin{algorithm}[t]
\caption{Dual Half-Space Projected Gradient (DHSPG)}
\label{alg:main.dhspg}
\begin{algorithmic}[1]
\State \textbf{Input:} initial variable $\bm{x}_0\in\mathbb{R}^n$, initial learning rate $\alpha_0$, warm-up steps $T$, target group sparsity $K$, group partition $\mathcal{G}_\text{prune}$, and the optimization variant $\mathcal{OPT}$.
\State Warm up $T$ steps via the selected $\mathcal{OPT}$.\label{line:warm_up}
\State Separate $\mathcal{G}_\text{redundant}$ and $\mathcal{G}_\text{important}$ given $\mathcal{G}_\text{prune}$ and $K$.\label{line:partition_groups}
\For {$t=T, T+1,T+2,\cdots, $}
	\State Compute gradient estimate $\Grad f(\bm{x}_t)$ based on $\mathcal{OPT}$.\label{line:compute_gradient_estimate}
	\State Update $[\bm{x}_{t+1}]_{\mathcal{G}_\text{important}}$ as 
	$[\bm{x}_t-\alpha_t \Grad f(\bm{x}_t)]_{\mathcal{G}_\text{important}}.
	$
	\State Select proper $\lambda_g$ for $g\in\mathcal{G}_\text{redundant}$.
	\State Compute $[\tilde{\bm{x}}_{t+1}]_{\mathcal{G}_\text{redundant}}$ via subgradient descent of $\psi$.
	\label{line:compute_trial_iterate}
    \State Perform Half-Space projection over $[\tilde{\bm{x}}_{t+1}]_{\mathcal{G}_\text{redundant}}$.
	\label{line:half_space_projection}
 \State Update $[\bm{x}_{t+1}]_{\mathcal{G}_\text{redundant}}\gets [\tilde{\bm{x}}_{t+1}]_{\mathcal{G}_\text{redundant}}$.
	\State Update $\alpha_{t+1}$.
\EndFor
\State \textbf{Return} the final iterate $\bm{x}^*_{\dhspg}$.
\end{algorithmic}
\end{algorithm}

\noindent
\textbf{Initiation and Warm-Up.} Initially, users are required to configure an optimization variant for gradient estimation (with supported options including SGD, Adam, and AdamW). This requirement stems from the anticipation that users might already possess a well-established training pipeline. In such pipelines, the baseline optimization consistently yields high performance for the full model. Therefore, we recommend users to select the same optimization variant as their existing setup, promoting generality and facilitating ease of integration. Afterwards, \dhspg{} undergoes $T$ steps of a warming-up process. This stage gathers gradient signals, which are instrumental in distinguishing between redundant and important variable groups. 

\noindent
\textbf{Redundant Identification.} After the warm-up phase, we classify the groups $\mathcal{G}_\text{prune}$ into redundant groups $\mathcal{G}_\text{redundant}$ and important groups $\mathcal{G}_\text{important}\gets\mathcal{G}_\text{prune}/ \mathcal{G}_\text{redundant}$. To construct this partition, various criteria based on salience scores can be applied. One could use the cosine similarity $\cos{(\theta_g)}$ between the projection direction $-[\bm{x}]_g$ and the negative gradient or its estimation $-[\Grad f(\bm{x})]_g$. Higher cos-similarity over $g\in \mathcal{G}_\text{PZIG}$ indicates that projecting the group of variables in $g$ onto zeros is more likely to make progress to the optimality of $f$ (considering the descent direction from the perspective of optimization). Therefore, we identify $\mathcal{G}_\text{redundant}$ by selecting the PZIGs with bottom-$K$ least salience scores and $\mathcal{G}_\text{important}$ as its complementary as~(\ref{eq:redundant_group_partition_pruning}),
\begin{equation}\label{eq:redundant_group_partition_pruning}
\mathcal{G}_\text{redundant}=(\text{Bottom-K})\mathop{\text{argmax}}_{g\in\mathcal{G}_\text{PZIG}}\text{salience-score}(g)\ \text{and}\ \mathcal{G}_\text{important}=\mathcal{G}_\text{prune}/  \mathcal{G}_\text{redundant}.
\vspace{-2mm}
\end{equation}

\begin{wrapfigure}{r}{0.4\textwidth}
\vspace{-4mm}
\includegraphics[width=0.9\linewidth]{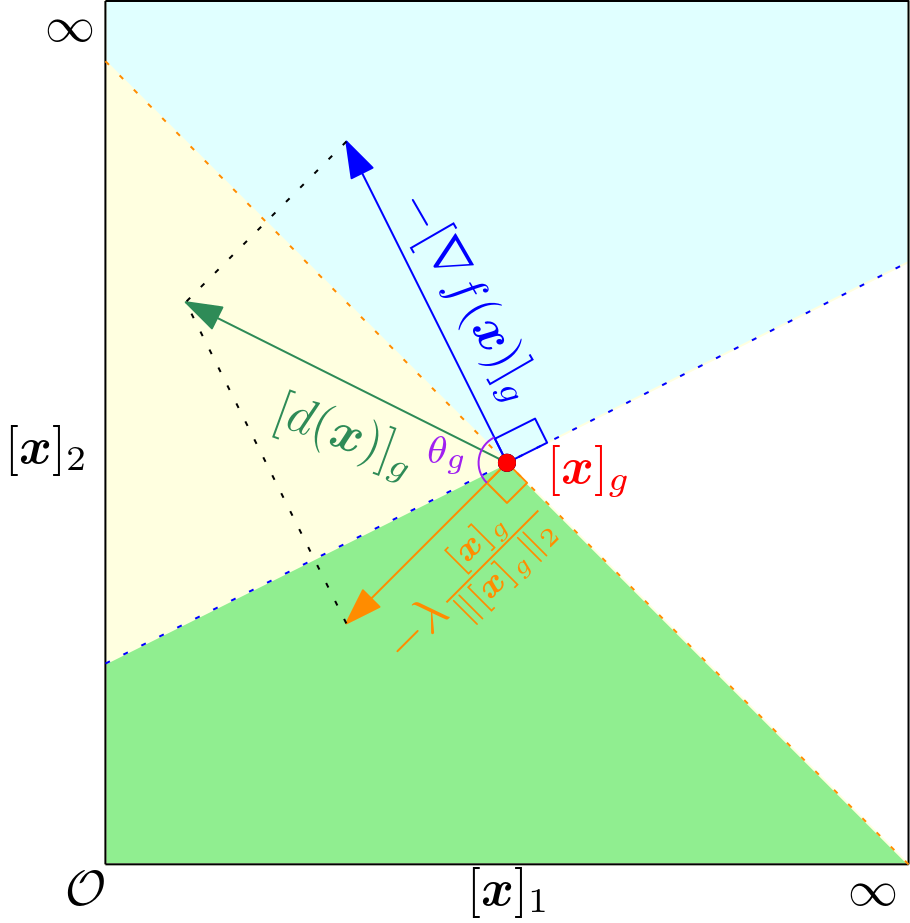}
\vspace{-3mm}
\caption{ Search direction in DHSPG.}
\vspace{-5mm}
\label{fig:search_direction_dhspg}
\end{wrapfigure}
\noindent
\textbf{Hybrid Training.} We then propose a \textit{hybrid} training schema that applies distinct update mechanisms to important and redundant variable groups, diverging from the standard optimization approaches where the same update schema is uniformly applied to all variables. 
For the important variable groups $\mathcal{G}_\text{important}$, we seek to achieve competitive performance, thereby adopt gradient descent using the selected gradient estimation method like Adam~\citep{kingma2014adam}, \ie, $[\bm{x}_{t+1}]_{\mathcal{G}_\text{important}}\gets [\bm{x}_t-\alpha_t \Grad f(\bm{x}_t)]_{\mathcal{G}_\text{important}}$.
For the redundant groups $\mathcal{G}_\text{redundant}$, our goal is to project these redundant variables towards zero for removing the corresponding pruning minimally removal structures. However, direct projection can disrupt progress towards the optimum. This issue becomes particularly severe in applications like Large Language Models (LLMs)~\citep{ding2023efficiency}, where redundant groups may still hold substantial knowledge~\citep{chen2023lorashear}. Direct projection risks entirely losing this knowledge, which could be challenging to regain. To address this, we advocate for a progressive approach that gradually diminishes their magnitudes and projects them onto zero without impairing the objective to the largest extent. This strategy is termed as \textit{inherent knowledge transfer}. Then the update over the redundant groups is formulated as a non-constrained subproblem as \eqref{prob.sub_penaly_group}.
\begin{equation}\label{prob.sub_penaly_group}
\minimize{[\bm{x}]_{\mathcal{G}_\text{redundant}}}\ \psi([\bm{x}]_{\mathcal{G}_\text{redundant}}):=f\left([\bm{x}]_{\mathcal{G}_\text{redundant}}\right)+\sum_{g\in\mathcal{G}_\text{redundant}}\lambda_g \norm{[\bm{x}]_{g}}_2,
\end{equation} 
where $\lambda_g$ is a group-specific regularization coefficient and needs to be dedicately chosen to guarantee the decrease of both the variable magnitude for $g$ as well as the objective $f$. In particular, we compute a negative subgradient of $\psi$ as the search direction $[\bm{d}(\bm{x})]_{\mathcal{G}_\text{redundant}}:=-[\Grad f(\bm{x})]_{\mathcal{G}_\text{redundant}}-\sum_{g\in\mathcal{G}_\text{redundant}}\lambda_g{[\bm{x}]_g/\max\{\norm{[\bm{x}]_g}_2}, \tau\}$ with $\tau$ as a safeguard constant. 
To ensure $\bm{d}(\bm{x})$ as a descent direction for both $f$ and $\norm{\bm{x}}_2$, $[\bm{d}(\bm{x})]_{g}$ needs to fall into the intersection between the \textit{dual half-spaces} with normal directions as $-[\Grad f]_g$ and $-[\bm{x}]_g$ for any $g\in\mathcal{G}_\text{redundant}$ as shown in Figure~\ref{fig:search_direction_dhspg}. In other words, $[\bm{d}(\bm{x})]_g^\top [-\Grad f(\bm{x})]_g$ and $[\bm{d}(\bm{x})]_g^\top [-\bm{x}]_g$ are greater than 0.
It further indicates that $\lambda_g$ locates in the interval 
$(\lambda_{\text{min},g}, \lambda_{\text{max},g}):= \left(-\cos(\theta_g)\norm{[\Grad f(\bm{x})]_g}_2, -\frac{\norm{[\Grad f(\bm{x})]_g}_2}{\cos{(\theta_g)}}\right)$
if $\cos{(\theta_g)}<0$ otherwise can be an arbitrary positive constant. Such $\lambda_g$ brings the decrease of both the objective and the variable magnitude. We then compute a trial iterate $[\tilde{\bm{x}}_{t+1}]_g\gets[\bm{x}_t-\alpha_t\bm{d}(\bm{x}_t)]_g$ via the subgradient descent of $\psi$ as line~\ref{line:compute_trial_iterate}. The trial iterate is fed into the Half-Space projector \citep{chen2021oto,dai2023adaptive} which outperforms proximal operators to yield group sparsity more productively without deteriorating the objective as line~\ref{line:half_space_projection}. In the end, we return the final or the best iterate that reaches target group sparsity $K$ and high-performance as the solution $\bm{x}_\text{DHSPG}^*$. 

\subsection{Automatic Structurally Pruned Network Construction}\label{sec.sub_network_pruning}

\begin{wrapfigure}{r}{0.55\textwidth}
\begin{minipage}{0.98\linewidth}
\vspace{-3mm}
\includegraphics[width=0.9\linewidth]{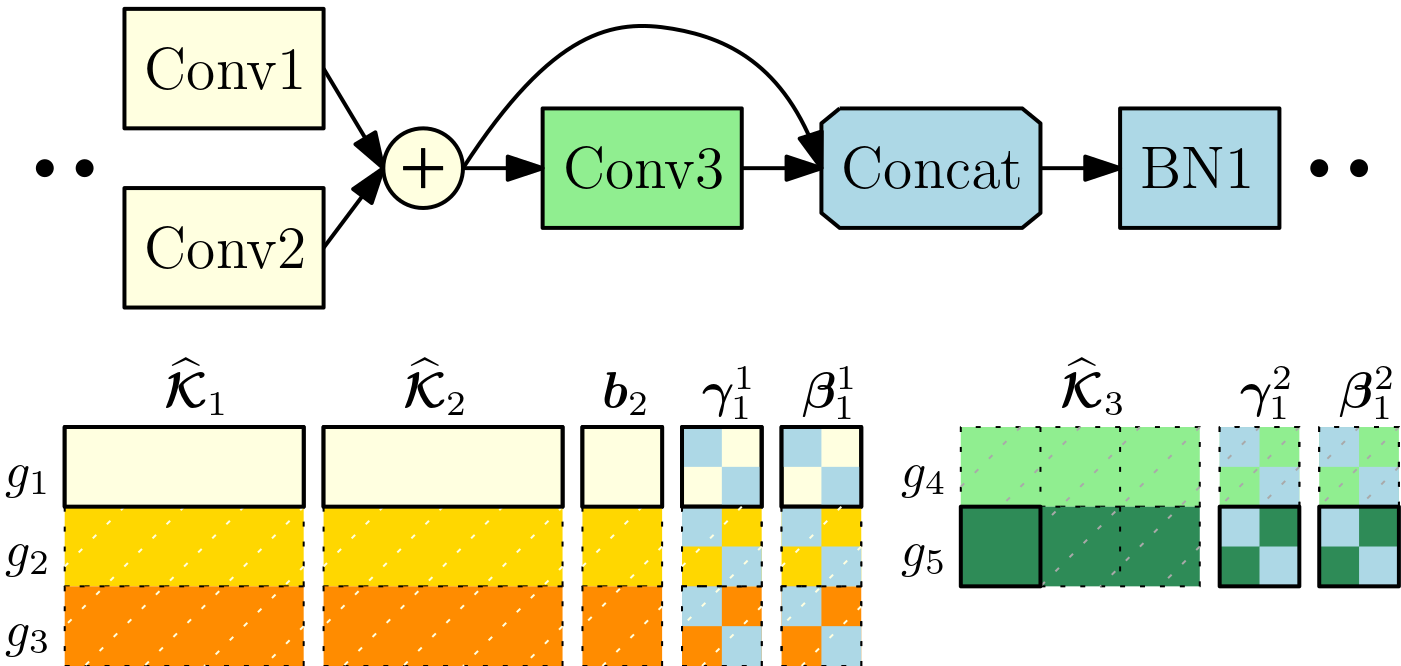}
\caption{Automated pruned model construction. $\mathcal{G}=\{g_1, g_2, \cdots, g_5\}$ and $[\bm{x_{\dhspg{}}}^*]_{g_2\cup g_3\cup g_4}=\bm{0}$.}
\label{fig:compressed_model_construction}
\end{minipage}
\vspace{-5mm}
\end{wrapfigure}
Ultimately, having derived the solution $\bm{x}^*_{\dhspg{}}$ that achieves both high performance and group sparsity, we now move to the \textit{automated} construction of a compact model, which is a \textit{manual} step with unavoidable substantial engineering efforts in most of structured pruning works. From an engineering perspective, in OTOv2~\citep{chen2023otov2}, the pruned model is produced in ONNX~\citep{onnxruntime} format, which is not conveniently for the further usage such as quantization if needed. In \algacro{}, we have made a significant engineering achievement to re-design, re-factorize and re-develop the whole library \textbf{to produce the pruned model directly in Pytorch format}, which greatly enhance the generality, flexibility, and coverage. From the algorithmic view, we traverse all vertices with trainable parameters, then remove the structures in accordance with PZIGs being zero, such as the dotted rows of $\widehat{\bm{\mathcal{K}}}_1, \widehat{\bm{\mathcal{K}}}_2, \widehat{\bm{\mathcal{K}}}_3$ and scalars of $\bm{b}_2, \bm{\gamma}_1, \bm{\beta}_1$ as illustrated in Figure~\ref{fig:compressed_model_construction}. Next, we erase the redundant parameters that affiliate with the removed structures of their incoming stem vertices to keep the operations valid, \eg, the second and third channels in $g_5$ are removed though $g_5$ is not zero. The automated algorithm is promptly complete in linear time via performing two passes of depth-first-search and manipulating parameters to produce a more compact model $\mathcal{M}^*_\text{prune}$. Based on the property of PZIGs, $\mathcal{M}^*_\text{prune}$ returns the same inference outputs as the full $\mathcal{M}$ parameterized as $\bm{x}^*_{\dhspg{}}$, thus no further fine-tuning is necessary.

\section{\algacro{} for Erasing Operator}\label{sec.erasing_mode}

The second fresh application domain of \algacro{} is the erasing mode, which aims at training a general DNN and automatically finding an optimal sub-network via erasing redundant operators entirely. The resulting sub-network is not only high-performing but also has a remarkably compact architecture, making it well-suited for various deployment environments. The erasing mode is closely related to a popular sub-realm in neural architecture search (NAS) to find an optimal sub-network given a network that covers all candidate operations and connections. Compared with these NAS methods~\citep{liu2018darts}, \algacro{} resolves one of their main pain-points that requires significant handcraftness and engineering efforts to determine the architecture specific search spaces. The erasing mode in \algacro{} provides an end-to-end automated solution, dramatically reduces the necessity for human intervention, and is compatible with a wide range of DNNs and applications. 

\paragraph{New Challenges Compared to Pruning Mode.} The erasing mode operates orthogonally to the pruning mode in Section~\ref{sec.pruning_mode} which preserves all operators yet slims them.  Such major difference poses several new challenges to the erasing mode. First of all, the erasing dependency across vertices differs from the pruning dependency, requiring a distinct dependency graph analysis to discover the erasing minimally removal structures. Secondly, the hierarchy among the discovered erasing minimally removal structures needs to be additionally taken account, which is crucial to ensure the validity of the resulting sub-networks, yet discarded in the pruning mode. Thirdly, topological structure of sub-network in the erasing mode is dramatically changed, which brings significant challenges into the sub-network construction component. 
As a result, 
each component in the erasing mode necessitates \underline{\textit{distinct \textbf{algorithmic design} and \textbf{engineering developments}}} to the pruning mode. This makes it not only a substantial extension to a new application domain but also represents a significant dive in terms of algorithmic innovation, distinguishing it from the existing works.


\subsection{Automatic Erasing Search Space Generation for Erasing Mode}\label{sec.search_space_erasing}

\begin{figure}[t]
    \centering
    \begin{subfigure}{\linewidth}
    	\centering
    \includegraphics[width=0.9\linewidth]{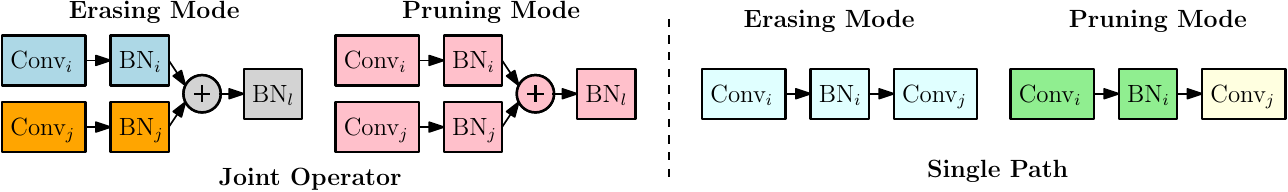}
    \caption{Erasing dependency versus pruning dependency.}\label{fig:diff_dependency}
    	\vspace{1mm}
    	\label{fig:1a}		
    \end{subfigure}
    \begin{subfigure}{\linewidth}
    	\centering
    	\includegraphics[width=\linewidth]{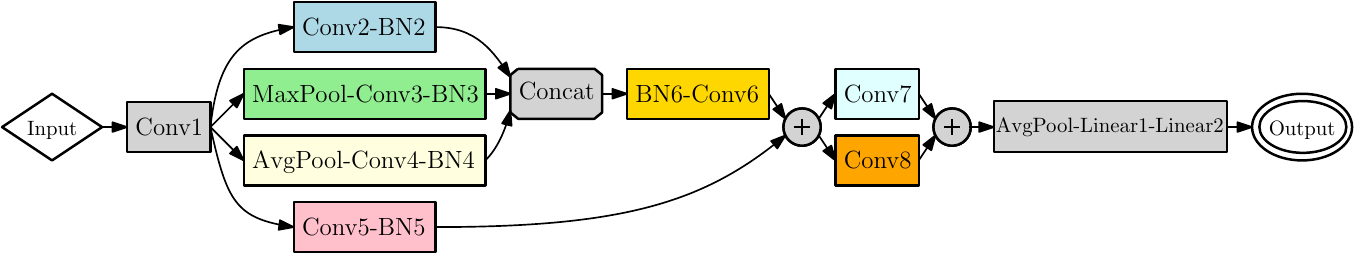}
    	\caption{Dependency graph for erasing operator.}\label{fig:demonet_dependency_erasing}
    	\vspace{1mm}
    	\label{fig:1a}		
    \end{subfigure}
    \begin{subfigure}{\linewidth}
    	\centering
    	\includegraphics[width=\linewidth]{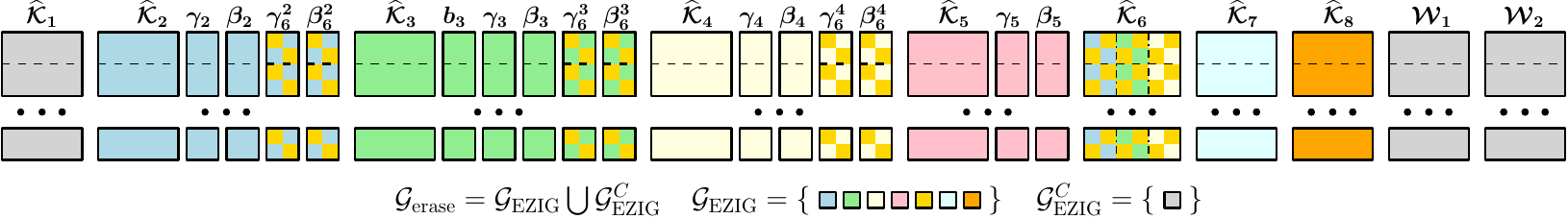}
    	\caption{Trainable variable partition.}\label{fig:erasing_zig}
    	\vspace{1mm}
    	\label{fig:1a}		
    \end{subfigure}
    \caption{
    Automated Search Space Generation. Given the \demonet{} to be trained and compressed in the erasing mode, (a) the constructed dependency graph for erasing mode; and (b) the trainable variable partition, where $\mathcal{G}_s$ represents the {variable groups} corresponding to removal structures. $\bm{\widehat{\mathcal{K}}}_i$ and $\bm{b}_i$ are the flatten filter matrix and bias vector for \texttt{Conv-i}, respectively. $\bm{\gamma}_i$ and $\bm{\beta}_i$ are the weight and bias vectors for \texttt{BN-i}. $\bm{{\mathcal{W}}}_i$ is the weight matrix for \texttt{Linear-i}. The columns of $\bm{\widehat{\mathcal{K}}}_6$ are marked in accordance to its incoming segments.
    }
    \label{fig:automatic_search_space_construction} 
    \vspace{-3mm}
\end{figure}

Given a pre-defined DNN $\mathcal{M}$, as outlined in Algorithm~\ref{alg:main.outline}, the initial step of the erasing mode in \algacro{} is to automatically generate a search space which is defined as a set of minimally removal structures for the erasing purpose (see Definition~\ref{def:search_space}). The automated generation of erasing search space involves two main phases. The first phase explores the trace graph of the DNN $\mathcal{M}$ and establishes a erasing dependency graph. This graph is \textit{notably different} from the pruning dependency graph discussed in Section~\ref{sec.search_space_pruning} due to the distinct dependent relations in separate compression modes as presented in Figure~\ref{fig:diff_dependency}.
The second phase leverages the erasing dependency graph to find out erasing minimally removal structures, then partitions their trainable variables to a set of erasing zero-invariant groups (EZIGs) and their unsearchable complementary. For intuitive illustrations, we still use the \demonet{} depicted as Figure~\ref{fig:demonet_original} to elaborate the erasing search space generation.

\paragraph{Erasing Dependency versus Pruning Dependency.} The erasing dependency is distinct from the pruning dependency. As drawn in Figure~\ref{fig:diff_dependency}, the same operators may behave distinctly regarding dependency in different modes. For example, in pruning mode, the joint operators, \eg, \texttt{Add} and \texttt{Mul}, introduce dependency across their incoming stem vertices to be pruned together, whereas in the erasing mode, their incoming stem vertices could operate independently. Another difference is in the treatment of consecutive stem vertices along the single path. In the pruning mode, these vertices need to form separate node groups. In contrast, during the erasing mode, they must be considered as a unified group. This is because erasing any of these vertices would result in the remaining ones being disconnected from either the input or the output of the DNN, thereby rendering the network invalid.

\paragraph{Erasing Dependency Graph.} 

\begin{algorithm}[h]
	\caption{Automated Erasing Search Space Generation.}
	\label{alg:search_space_erasing}
	\begin{algorithmic}[1]
		\State \textbf{Input:} A neural network $\mathcal{M}$ to be trained and compressed via erasing operators. 
        \State \textit{\textbf{Erasing dependency graph construction.}}\label{line:erasing_dependency_start}
        \State Construct the trace graph $(\mathcal{E}, \mathcal{V})$ of $\mathcal{M}$.\label{line:vertices_edges_trace_graph}
        \State Initialize an empty graph $(\mathcal{V}_\text{erase},\mathcal{E}_\text{erase})$.
        \State Initialize queue $\mathcal{Q}\gets \{\mathcal{S}(v): v\in\mathcal{V}\text{ is adjacent to the input of trace graph}\}$.\label{line.initialize_queue}
        \While{$\mathcal{Q}\neq \emptyset$}
            \State Dequeue the head segment $\mathcal{S}$ from $\mathcal{Q}$.
            \State Grow $\mathcal{S}$ in the depth-first manner till meet either joint vertex or multi-outgoing vertex $\hat{v}$.\label{line.meet_endpoint_vertex}
            \State Add segments into $\mathcal{V}_\text{erase}$ and connections into $\mathcal{E}_\text{erase}$.
            \State Enqueue new segments into the tail of $\mathcal{Q}$ if $\hat{v}$ has outgoing vertices.\label{line.enqueue_new_elements}\label{line:erasing_dependency_end}
        \EndWhile
        \State \textit{\textbf{Find erasing minimal removal structures.}}
        \State Get the incoming vertices $\widehat{\mathcal{V}}$ for joint vertices in the $(\mathcal{V}_\text{erase}, \mathcal{E}_\text{erase})$.\label{line.incoming_vertices_of_joint}
        \State Group the trainable variables in the vertex $v\in\widehat{\mathcal{V}}$ as $g_v$.\label{line:group_variables_vertex_zig}
        \State Form $\mathcal{G}_\text{EZIG}$ as the union of the above groups, \ie, $\mathcal{G}_\text{EZIG}\gets\{g_v: v\in \widehat{\mathcal{V}}\}$.
        \label{line.form_zigs}
        \State Form $\mathcal{G}_\text{EZIG}^C$ as the union of the trainable variables in the remaining vertices. 
	\State \textbf{Return} variable partition $\mathcal{G}=\mathcal{G}_\text{EZIG}\cup \mathcal{G}_\text{EZIG}^C$ and erasing dependency graph $(\mathcal{V}_\text{erase},\mathcal{E}_\text{erase})$.
	\end{algorithmic}
\end{algorithm}

To track the dependency in erasing mode, we design a novel graph algorithm stated as line~\ref{line:erasing_dependency_start}-\ref{line:erasing_dependency_end} in Algorithm~\ref{alg:search_space_erasing}. We start from the trace graph $(\mathcal{V}, \mathcal{E})$ of the target DNN $\mathcal{M}$  as Figure~\ref{fig:demonet_original} and line~\ref{line:vertices_edges_trace_graph}, 
then analyze it to create the erasing dependency graph $(\mathcal{V}_\text{erase}, \mathcal{E}_\text{erase})$, wherein each vertex in $\mathcal{V}_\text{erase}$ serves as a potential erasing minimally removal structure candidate.

To proceed, we use a queue container $\mathcal{Q}$ to track the erasing minimally removal structure candidates (line \ref{line.initialize_queue}). The initial elements of this queue are the vertices that are directly adjacent to the input of $\mathcal{M}$, such as \texttt{Conv1}. We then traverse the graph in the breadth-first manner, iteratively growing each element (segment)  $\mathcal{S}$ in the queue until a valid minimally removal structure candidate is formed. The growth of each candidate follows the depth-first search to recursively expand $\mathcal{S}$ until the current vertices are considered as endpoints. An endpoint vertex is determined by whether it is a joint vertex or has multiple outgoing vertices, as indicated in line~\ref{line.meet_endpoint_vertex}. Intuitively, a joint vertex has multiple inputs, which means that the DNN may be still valid after removing the current segment. This suggests that the current segment may be removable. On the other hand, a vertex with multiple outgoing neighbors implies that removing the current segment may cause some of its children to miss the input tensor. For instance, removing \texttt{Conv1} would cause \texttt{Conv2}, \texttt{MaxPool} and \texttt{AvgPool} to become invalid due to the absence of input in Figure~\ref{fig:demonet_original}. Therefore, it is risky to remove such candidates. Once the segment $\mathcal{S}$ has been grown, new candidates are initialized as the outgoing vertices of the endpoint and added into the container $\mathcal{Q}$ (line~\ref{line.enqueue_new_elements}). Such procedure is repeated until the end of traversal and returns the erasing dependency graph $(\mathcal{V}_\text{erase}, \mathcal{E}_\text{erase})$ as Figure~\ref{fig:demonet_dependency_erasing}.

\paragraph{Erasing Zero-Invariant Group Partition.} We now identify the erasing minimally removal structures upon $(\mathcal{V}_\text{erase}, \mathcal{E}_\text{erase})$ to form the erasing search space. The qualified instances are the vertices in $\mathcal{V}_\text{erase}$ that have trainable variables and all of their outgoing vertices are joint vertices. This is because a joint vertex has multiple inputs and remains valid even after removing some of its incoming structures, as indicated in line~\ref{line.incoming_vertices_of_joint} in Algorithm~\ref{alg:search_space_erasing}. Consequently, their trainable variables are grouped together into erasing zero invariant groups $\mathcal{G}_\text{EZIG}$ (line \ref{line:group_variables_vertex_zig}-\ref{line.form_zigs} and Figure~\ref{fig:erasing_zig}). The remaining vertices are considered as either unremovable or belonging to a large removal structure, which trainable variables are grouped into the $\mathcal{G}_\text{EZIG}^C$ (the complementary to $\mathcal{G}_\text{EZIG}$). As a result, for the given DNN $\mathcal{M}$, all its trainable variables are encompassed by the union $\mathcal{G}_\text{erase}\gets\mathcal{G}_\text{EZIG}\cup \mathcal{G}_\text{EZIG}^C$. The corresponding erasing minimally removal structures to the variable groups $\mathcal{G}_\text{EZIG}$ constitute the erasing search space.

\subsection{Hiearachical Half-Space Projected Gradient (\hhspg{})}\label{sec.hhspg}

Given the target DNN $\mathcal{M}$ and its variable group partition $\mathcal{G}_\text{erase}=\mathcal{G}_\text{EZIG}\cup \mathcal{G}_\text{EZIG}^C$, the next is to jointly search for a valid sub-network $\mathcal{M}^*$ that exhibits the most significant performance and train it to high performance. Searching a sub-network is equivalent to identifying the redundant EZIGs in $\mathcal{G}_\text{EZIG}$, erasing the corresponding operators, and ensuring the remaining network still valid. Training the sub-network becomes optimizing over the remaining important groups in $\mathcal{G}_\text{erase}$ to achieve high performance. Structured sparsity problem is a natural choice to accomplish both tasks. However, compared to the pruning mode, the erasing mode needs to additionally consider the hierarchy across the erasing minimally removal structures, leading to a more challenging a \underline{\textit{hierarchical}} structured sparsity problem formulated as~\eqref{prob.main.erasing}, which has not been addressed for the deep learning application. 
\begin{equation}\label{prob.main.erasing}
\minimize{\bm{x}\in \mathbb{R}^n}\ f(\bm{x}),\ \ \text{s.t.} \ \text{Cardinality}(\mathcal{G}^{0})=K,\ \text{and}\  (\mathcal{V}_\text{erase}/\mathcal{V}_{\mathcal{G}^{0}}, \mathcal{E}_\text{erase}/\mathcal{E}_{\mathcal{G}^{0}})\ \text{is valid},
\end{equation}
\noindent
where $f$ is the prescribed loss function, $\mathcal{G}^{=0}:=\{g\in\mathcal{G}_\text{EZIG}| [\bm{x}]_g=0\}$ is the set of zero groups in $\mathcal{G}_\text{EZIG}$, which cardinality measures its size. $K$ is the target hierarchical group sparsity, indicating the number of erasing minimally removal structures that should be identified as redundant. A larger $K$ dictates a higher sparsity level that produces a more compact sub-network with fewer FLOPs and parameters. The trainable variables in the redundant structures are projected onto zero, while the trainable variables in the important structures are preserved as non-zero and optimized for high performance.   $(\mathcal{V}_\text{erase}/\mathcal{V}_{\mathcal{G}^{0}}, \mathcal{E}_\text{erase}/\mathcal{E}_{\mathcal{G}^{0}})$ refers to the sub-graph removing vertices and edges according to zero groups $\mathcal{G}^0$. It being valid requires the zero groups distributed obeying the hierarchy of the erasing dependency graph to ensure the resulting sub-network functions correctly.

\begin{algorithm}[t]
\caption{Hierarchical Half-Space Projected Gradient (\hhspg)}
\label{alg:main.h2spg}
\begin{algorithmic}[1]
\State \textbf{Input:} initial variable $\bm{x}_0\in\mathbb{R}^n$, initial learning rate $\alpha_0$, target group sparsity $K$, warm-up steps $T$, optimization variant $\mathcal{OPT}$, erasing dependency graph $(\mathcal{V}_\text{erase}, \mathcal{E}_\text{erase})$ and group partition $\mathcal{G}_\text{erase}=\mathcal{G}_\text{EZIG}\cup \mathcal{G}_\text{EZIG}^C$.
\State \textit{\textbf{Hierarchical Search Phase.}}
\State Initialize the group set for redundant erasing minimally removal structures $\mathcal{G}_\text{redundant}\gets \emptyset$. 
\State Initialize remaining sub-graph $(\widehat{\mathcal{V}}, \widehat{\mathcal{E}})\gets (\mathcal{V}_\text{erase}, \mathcal{E}_\text{erase})$.
\State Calculate the saliency score via modular proxy for each $g\in \mathcal{G}_\text{EZIG}$ and sort them.  \label{h2spg.salience_score_calculation}
\For {$g\in \mathcal{G}_\text{EZIG}$ ordered by saliency scores ascendingly}\label{line.check_start}
    \State Find the vertex $v_g$ for $g$ and the adjacent edges $\mathcal{E}_g$. \label{line.find_vertice_edges}
    \If {$(\widehat{\mathcal{V}}/\{v_g\}, \widehat{\mathcal{E}}/ \mathcal{E}_g)$ is valid and $|\mathcal{G}_\text{redundant}|< K$}\label{line.check_valid_remain_graph}
        \State Update $\mathcal{G}_\text{redundant}\gets \mathcal{G}_\text{redundant}\cup \{g\}$.
        \State Update $(\widehat{\mathcal{V}},\widehat{\mathcal{E}})\gets(\widehat{\mathcal{V}}/\{v_g\}, \widehat{\mathcal{E}}/ \mathcal{E}_g)$. \label{line.check_end}
    \EndIf
\EndFor
\State \textit{\textbf{Hybrid Training Phase.}}
\For {$t=0, 1,\cdots, $}
	\State Compute gradient estimate $\Grad f(\bm{x}_t)$ by the selected $\mathcal{OPT}$.\label{line.compute_gradient_estimate}
	\State Update $[\bm{x}_{t+1}]_{\mathcal{G}_\text{redundant}^C}$ as 
	$[\bm{x}_t-\alpha_t \Grad f(\bm{x}_t)]_{\mathcal{G}_\text{redundant}^C}.
	$\label{line.update_important_groups}
	\State Perform Half-Space projection over $[\bm{x}_{t}]_{\mathcal{G}_\text{redundant}}$.
	\label{line:half_space_projection}
\EndFor
\State \textbf{Return} the final or the best iterate as $\bm{x}^*_{\hhspg{}}$.
\end{algorithmic}
\end{algorithm}

\begin{figure}[t]
\centering
\includegraphics[width=\linewidth]{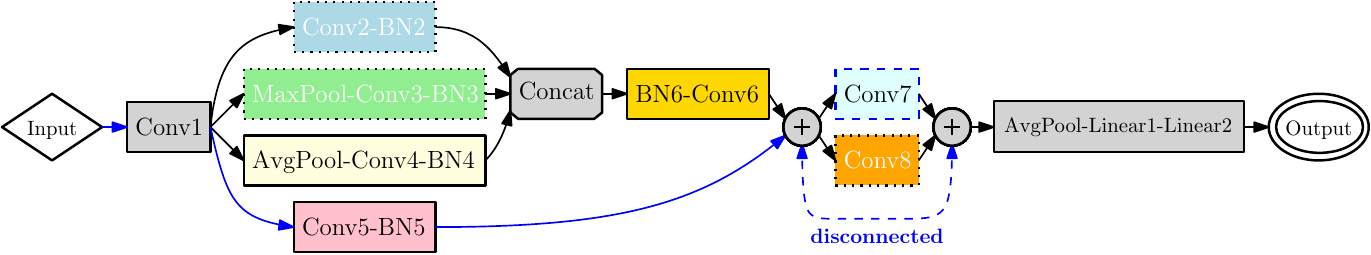}
\caption{Check validness of redundant candidates. Target group sparsity $K=3$. \texttt{Conv7} has smaller salience score than \texttt{Conv2-BN2}. Dotted vertices are marked as redundant candidates.} \label{fig:check_validaness_group_sparsity} 
\vspace{-5mm}
\end{figure}

\paragraph{Related Sparse Optimizers.} Problem~\eqref{prob.main.erasing} is difficult to solve due to the non-differential and non-convex sparsity constraint and the graph validity constraint. Existing sparse optimizers such as HSPG~\citep{chen2020half,dai2023adaptive} and proximal methods~\citep{deleu2021structured,klopfenstein2023local} overlook the architecture evolution and hierarchy during the sparsity exploration, which is crucial to (\ref{prob.main.erasing}). In fact, they are mainly applied for the pruning tasks, wherein operations are preserved yet become slimmer. Consequently, employing them onto (\ref{prob.main.erasing}) usually produces invalid sub-networks. 

\paragraph{Outline of \hhspg{}.} To effectively solve problem (\ref{prob.main.erasing}), we propose a novel \hhspg{} to consider the hierarchy and ensure the validness of graph architecture after removing redundant vertices during the optimization process. 
To the best of our knowledge, \hhspg{} is the first optimizer that successfully solves such hierarchical structured sparsity problem, which outline is stated in Algorithm~\ref{alg:main.h2spg}.

H2SPG is a hybrid multi-phase optimizer, distinguished by its dedicated designs catering to the hierarchical constraint, positioning it significantly apart from its non-hierarchical counterparts within the HSPG sparse optimizer family~\citep{chen2020half,chen2023otov2,dai2023adaptive}. Initially, \hhspg{} classifies variable groups into important and redundant segments through a novel \textit{hierarchical search phase}. The hierarchical search phase considers the topology of erasing dependency graph $(\mathcal{V}_\text{erase}, \mathcal{E}_\text{erase})$ to ensure the validness of the resulting sub-network. Subsequently, it applies the hybrid training phase proposed in Section~\ref{sec.dhspg} to employ separate updating mechanisms onto different segments to achieve a solution with both desired hierarchical group sparsity and high performance.

\paragraph{Hierarchical Search Phase.}  \hhspg{} first computes the saliency scores for each erasing zero-invariant groups (EZIGs) (line~\ref{h2spg.salience_score_calculation} in Algorithm~\ref{alg:main.h2spg}). The saliency score measures the importance of each erasing minimally removal structures to form an optimal sub-network. It design and calculation are modular to varying proxies, \eg, gradient-based proxies or training-free zero-shot proxies \citep{ming_zennas_iccv2021,chen2021neural,li2023zico} upon the need of downstream tasks. If fidelity is the main focus, the score should measure from the optimization perspective. If efficiency on hardware is the main focus, the score should favor more on hardware acceleration~\citep{li2023hardware,ghimire2022survey}. We by default use the gradient-based proxy due to its generality. In particular,  we first warm up all variables by conducting SGD or its variants. During the warm-up, a salience score of each group $g\in\mathcal{G}_\text{EZIG}$ is computed. Smaller salience score indicates the group exhibits less prediction power, thus may be redundant. By default, we consider both the cosine similarity between negative gradient $-[\Grad f(\bm{x})]_g$ and the projection direction $-[\bm{x}]_g$ as well as the average magnitude as \dhspg{} in Section~\ref{sec.dhspg}. The former one measures the approximate degradation of the objective function over the projection direction. 
The latter one measures the distance to the origin. 

The next is to form a set of redundant erasing minimally removal structure candidates $\mathcal{G}_\text{redundant}$ and ensures the validity of remaining DNN after erasing these candidates (line~\ref{line.check_start}-\ref{line.check_end} in Algorithm~\ref{alg:main.h2spg}). To proceed, we iterate each group in $\mathcal{G}_\text{EZIG}$ in the ascending order of salience scores. A remaining sub-graph $(\widehat{\mathcal{V}}, \widehat{\mathcal{E}})$ is constructed by iteratively removing the vertex of each group along with the corresponding adjacent edges from $(\mathcal{V}_\text{erase}, \mathcal{E}_\text{erase})$. The sanity check verifies whether the graph $(\widehat{\mathcal{V}}, \widehat{\mathcal{E}})$ is still connected after the erasion. If so, the variable group for the current vertex is added into $\mathcal{G}_\text{redundant}$; otherwise, the subsequent group is turned into considerations. As illustrated in Figure~\ref{fig:check_validaness_group_sparsity}, though \texttt{Conv7} has a smaller salience score than \texttt{Conv2-BN2}, \texttt{Conv2-BN2} is marked as potentially redundant but not \texttt{Conv7} since there is no path connecting the input and the output of the graph after removing \texttt{Conv7}. This mechanism largely guarantees that even if all redundant candidates are erased, the resulting sub-network is still functioning as normal. The complementary groups with higher salience scores are marked as important groups and form $\mathcal{G}_\text{important}\gets\mathcal{G}_\text{erase}/\mathcal{G}_\text{redundant}$.

\paragraph{Hybrid Training Phase.} \hhspg{} then engages into the hybrid training phase to produce desired hierarchical sparsity over $\mathcal{G}_\text{redundant}$ and optimize over $\mathcal{G}_\text{important}$ for pursuing excellent performance till the convergence. This phase has been described in Section~\ref{sec.dhspg},  and is briefly went through for completeness. In general, for the important groups $\mathcal{G}_\text{important}$, the vanilla SGD or its variant is employed to minimize the objective function (line~\ref{line.compute_gradient_estimate}-\ref{line.update_important_groups} in Algorithm~\ref{alg:main.h2spg}). For redundant group candidates in $\mathcal{G}_\text{redundant}$, a Half-Space projection step {introduced in \citep{chen2020half}} is proceeded to progressively yield sparsity without sacrificing the objective function to the largest extent. Finally, a high-performing solution $\bm{x}_\hhspg^*$ with desired hierarchical sparsity is returned. 

\begin{figure}[t]
    \centering
    \begin{subfigure}{\linewidth}
    	\centering
    	\includegraphics[width=\linewidth]{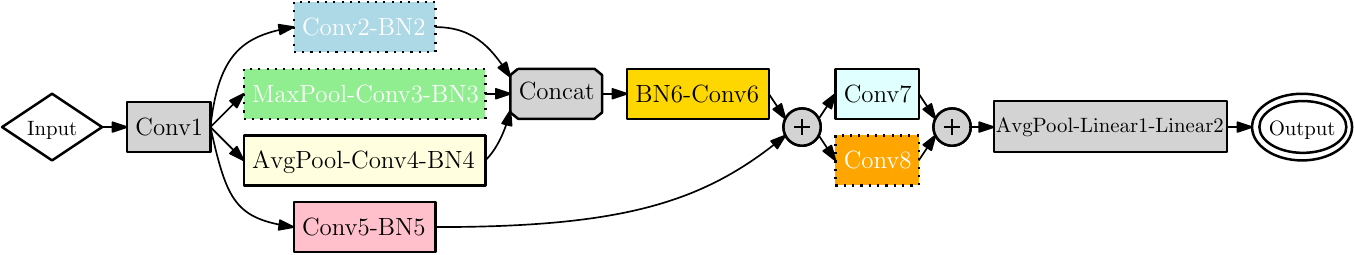}
    	\caption{Identified redundant structures.}\label{fig:redundant_structures}
    	\label{fig:1a}		
    \end{subfigure}
    \begin{subfigure}{0.45\linewidth}
    	\centering
    	\includegraphics[width=\linewidth]{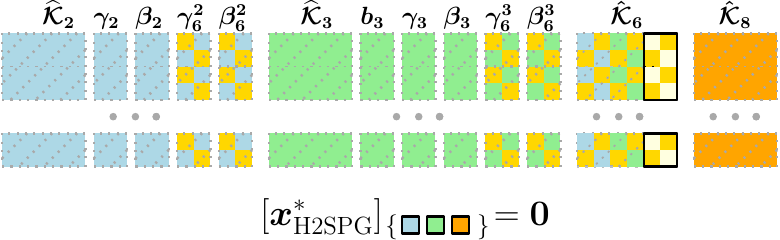}
    	\caption{Redundant erasing zero-invariant groups.}\label{fig:zero_groups}
    	\label{fig:1a}		
    \end{subfigure}
    \begin{subfigure}{0.54\linewidth}
    	\centering
    	\includegraphics[width=\linewidth]{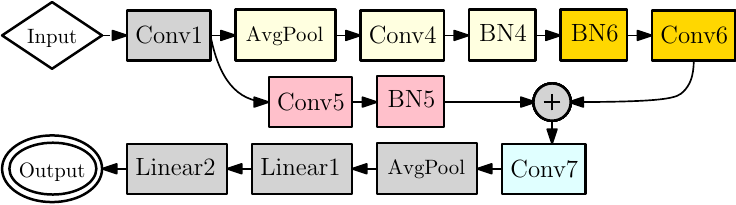}
    	\caption{Constructed sub-network.}\label{fig:constructured_sub_network}
    	\label{fig:1a}		
    \end{subfigure}
    \caption{Redundant erasing minimally removal structures and sub-network construction.}
    \label{fig:sub_network_construction}
    \vspace{-5mm}
\end{figure}

\subsection{Automatic Erased Network Construction}\label{sec.sub_network_erasing}

We finally construct a sub-network $\mathcal{M}^*_\text{erase}$ upon the given DNN $\mathcal{M}$ and the solution $\bm{x}_\hhspg^*$ by \hhspg{}. The solution $\bm{x}_\hhspg^*$ should attain desired target hierarchical group sparsity level and achieve high performance. As illustrated in Figure~\ref{fig:sub_network_construction}, we first traverse the graph to remove the vertices and the edges corresponding to the redundant erasing removal structures parameterized as zero. For example, \texttt{Conv2-BN2}, \texttt{MaxPool-Conv3-BN3} and \texttt{Conv8} are removed due to their variable groups being zero.  Then, we traverse the graph in the second pass to adjust the affiliated structures that are dependent on the removed vertices to keep the remaining operations valid, \eg, the first and second columns in $\bm{\widehat{\mathcal{K}}_6}$ are erased since its incoming vertices \texttt{Conv2-BN2} and \texttt{MaxPool-Conv3-BN3} have been removed (see Figure~\ref{fig:zero_groups}). Next, we recursively erase unnecessary vertices and isolated vertices. Isolated vertices refer to the vertices that have neither incoming nor outgoing vertices. Unnecessary vertices refer to the skippable operations, \eg, \texttt{Concat} and \texttt{Add} (between \texttt{Conv7} and \texttt{AvgPool}) become unnecessary. Ultimately, a compact sub-network $\mathcal{M}^*$ is constructed as shown in Figure~\ref{fig:constructured_sub_network}. Fine-tuning the constructed sub-network $\mathcal{M}^*_\text{erase}$ is optional and often not necessary.
\section{Numerical Experiments}

In this section, we demonstrate the autonomy, generality, and high-performance of the proposed \algacro{} in both pruning and erasing modes. We cover a wide range of benchmark DNN architectures, including Bert~\citep{NIPS2017_3f5ee243}, CARN~\citep{ahn2018fast}, ConvNeXt~\citep{liu2022convnet}, DemoNet, DARTS~\citep{liu2018darts}, DenseNet~\citep{huang2017densely}, RegNet~\citep{radosavovic2020designing}, ResNet~\citep{he2016deep}, StackedUnets~\citep{ronneberger2015u} and \vgg{}~\citep{simonyan2014very}. The applications are ranging from popular image classification, low-level computer vision, to natural language processing. \algacro{} proceeds end-to-end training and compression in separate modes along with the minimization of human manual efforts and achieves competitive even superior performance to the state-of-the-arts. 

\subsection{Library Implementation}

\algacro{} makes a significant leap in engineering achievements compared to its predecessor, OTOv2. A primary limitation of OTOv2 was its reliance on ONNX~\citep{onnxruntime}, which restricted the use of structurally pruned models, such as post-training or quantization if needed. \algacro{} addresses this issue by redesigning and refactoring the pruning mode to directly produce the pruned models in the Torch format. This enhancement markedly improves the generality and flexibility of \algacro{}. The end-users can now input a Torch-based DNN, train and compress it via \dhspg{}, and ultimately receive a pruned Torch model suitable for various applications, as presented in Figure~\ref{fig:overview}. The entire process is automated, and requires minimally manual adjustments to the original training pipeline. Furthermore, the erasing mode in \algacro{} explores a brand new application domain regarding automatically erasing redundant operators from DNNs. This mode is facing \textit{more significant engineering challenges} due to the dramatic topological transformation on the DNN architecture compared with the pruning mode. At present, we support generating the sub-network in ONNX format, and aim to push the engineering boundaries to produce Torch-based sub-networks in future iterations.

\subsection{Experimental Setup}

Given a full DNN, we  typically expect that the end-users should already have a stable training pipeline including an optimizer and a learning rate scheduler to reliably train the full DNN to high-performance. \algacro{} is designed to seamlessly plug into varied DNN training tasks. Consequently, we used the same gradient estimation / optimizer variant and learning rate scheduler as each baseline training pipeline into \dhspg{} and \hhspg{} throughout the below experiments. We recommend to identify redundancy at initialization stage yet after a fair enough warm-up steps. Therefore, the warm-up phase of \dhspg{} in the pruning mode is empirically set as about 1/10 of total training steps. For the erasing mode, \hhspg{} follows the existing gradient-based NAS works~\citep{liu2018darts} by performing 50 epochs for the hierarchical search phase. In the end, it is noteworthy that \algacro{} supports training from either scratch or a pre-trained checkpoint, \eg, Bert~\citep{NIPS2017_3f5ee243}. 

\subsection{Structurally Pruning Operators}

\begin{wraptable}{r}{0.6\textwidth}
\vspace{-4mm}
\scriptsize
\caption{\small Structurally pruning CARNx2.}
\vspace{-3mm}
\resizebox{\linewidth}{!}{
\begin{tabular}{l|c|c|c|ccc}
\Xhline{3\arrayrulewidth}
\multirow{2}{*}{Method} &  \multirow{2}{*}{Optimizer} &  \multirow{2}{*}{FLOPS} & \multirow{2}{*}{\# of Params} & \multicolumn{3}{c}{PSNR} \\
\cline{5-7}
 & & & & Set14 & B100 & Urban100 \\ \hline
Baseline & Adam & 100\% & 100\% & 33.5 & 32.1 & 31.5\\
\textbf{\algacro{} Pruning Mode} & \textbf{DHSPG}  & \textbf{24.3\%} & \textbf{24.1\%}  & \textbf{33.2}  & \textbf{31.9}  & \textbf{31.1} \\
\Xhline{3\arrayrulewidth}
\end{tabular}
}
\label{tab:pruning_carn}
\vspace{-4mm}
\end{wraptable}
\paragraph{CARNx2 on Super-Resolution.} We first select a popular architecture CARN~\citep{ahn2018fast} for the super-resolution task with the scaling factor as two, referring as CARNx2. We use benchmark DIV2K dataset~\citep{agustsson2017ntire} for training and Set14~\citep{zeyde2010single}, B100~\citep{martin2001database} and Urban100~\citep{huang2015single} datasets for evaluation. \algacro{} automatically discovers the pruning minimally removal structures and partitions the variables into PZIGs. Following~\citep{agustsson2017ntire}, we use Adam as the optimization variant into DHSPG which leverages both first and second order momentums to estimate gradient. Under the same learning rate scheduler and training steps as the baseline as well as a  target group sparsity as 50\%, \algacro{} could dramatically reduce about 76\% FLOPs and parameters with negligible PSNR degradation as presented in Table~\ref{tab:pruning_carn}.

\paragraph{\vggbn{} on \cifar{}.} We then focus on \vggbn{}, using one popular dataset \cifar{} \citep{Krizhevsky09}. \vggbn{} is a modified version of the original \vgg{}, incorporating a batch normalization layer after each convolutional layer. Similarly to CARNx2, \algacro{} automatically identifies the redundant pruning minimal removal structures, and simultaneously trains the important PZIGs to high performance from scratch without fine-tuning.  As the results shown in Table~\ref{table:vgg_cifar}, \algacro{} effectively structurally prunes \vggbn{}, maintains the baseline accuracy while using only 5.0\% of the parameters and 26.6\% of the FLOPs. It's noteworthy that, although SCP and RP achieve higher accuracy levels, they require significantly more computational resources, demanding 43\%-102\% more FLOPs compared to \algacro{}.

\begin{table}[t]
\centering
\caption{Structurally pruning \vggbn{} on \cifar{}. Convolutional layers are in bold.}
\label{table:vgg_cifar}
\vspace{-2mm}
\resizebox{\textwidth}{!}{
    \begin{tabular}{l|c|c|c|c|c}
        \Xhline{3\arrayrulewidth}
        Method & BN & Architecture & FLOPs & \# of Params &  Top-1 Acc. \\
        \hline
        Baseline & \boldcheckmark & \textbf{64-64-128-128-256-256-256-512-512-512-512-512-512}-512-512 & 100\% & 100\% & 93.2\% \\
        EC~\citep{li2016pruning} &  \boldcheckmark &
        \textbf{32-64-128-128-256-256-256-256-256-256-256-256-256}-512-512 &  65.8\% & 37.0\% & 93.1\% \\
        Hinge~\citep{li2020group} & \boldcheckmark & -- & 60.9\% & 20.0\% & 93.6\% \\ 
        SCP~\citep{kang2020operation} & \boldcheckmark & -- & 33.8\% & 7.0\% & 93.8\%\\
        RP~\citep{li2022revisiting} & \boldcheckmark & -- & 47.9\%  & 42.1\%  & \textbf{93.9\%} \\
        CPGCN~\citep{dichannel2022} & \boldcheckmark & -- & 26.9\%  & 6.9\%  & 93.1\% \\
        \textbf{\algacro{} Pruning Mode} & \boldcheckmark & \textbf{14-51-77-122-183-146-92-41-16-13-8-11-14}-107-183 & \textbf{26.6\%} & \textbf{5.0\%} & 93.4\%\\ 
        \hline
    \Xhline{3\arrayrulewidth} 
\end{tabular}}
\vspace{-5.5mm}
\end{table}

\paragraph{\densenetonetwoone{} on \cifarhundred{}.} We next consider \densenetonetwoone{}~\citep{huang2017densely} on the benchmark dataset \cifarhundred{} \citep{Krizhevsky09}.  As presented in Table~\ref{tab:densenet_convnext}, with target group sparsity as 70\%, the pruning mode in \algacro{} could train and compress from scratch, and dramatically reduce 79.2\% FLOPs with competitive top-1 accuracy to the baseline full model. 

\begin{table}[h]
    \scriptsize
    \centering
    \vspace{-1mm}
   \caption{Structurally Pruning DenseNet and ConvNeXt.}
   \vspace{-3mm}
	\begin{tabular}{l|c|c|c|c|c}
	\Xhline{3\arrayrulewidth}
        Method & Backend  & Dataset & FLOPs & \# of Params & Top-1 Acc. \\
        \hline
        Baseline & DenseNet121 & \cifarhundred{} & 100\% & 100\% & 77.0\% \\
        \textbf{\algacro{} Pruning Mode} &  DenseNet121 & \cifarhundred{} & \textbf{20.8\%} & \textbf{26.7\%} & \textbf{75.5\%}  \\ 
        \hdashline
        Baseline &  {ConvNeXt-Tiny} & ImageNet & 100\% & 100\% & 82.0\% \\
        \textbf{\algacro{} Pruning Mode} & {ConvNeXt-Tiny} & ImageNet & \textbf{52.8\%} & \textbf{54.2\%} & \textbf{81.1\%} \\
    \Xhline{3\arrayrulewidth}
	\end{tabular}
  \vspace{-5mm}
    \label{tab:densenet_convnext}
\end{table}

\paragraph{ConvNeXt-Tiny on \imagenet{}.} To validate the compliance of \algacro{} with popular training tricks such as TIMM~\citep{rw2019timm},  we employed the pruning mode onto ConvNeXt-Tiny for \imagenet{}, which baseline was trained underneath TIMM. The results reported in Table~\ref{tab:densenet_convnext} indicate that under the target group sparsity level as 50\%, the pruned  ConvNeXt-Tiny could reduced about 47.2\% FLOPs and 45.8\% parameters with only 0.9\% top-1 accuracy regression.

\paragraph{\resnetfifty{} on \imagenet{}.} 

\begin{wrapfigure}{r}{0.47\textwidth}
\vspace{-4mm}
\begin{minipage}{\linewidth}
\includegraphics[width=\linewidth]{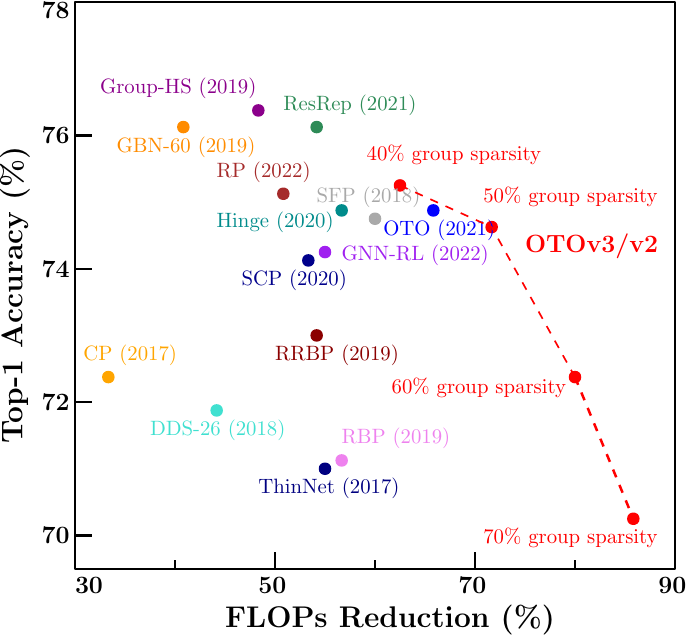}
    \caption{ResNet50 on ImageNet.}
    \label{fig:resnet50_imagenet}
\end{minipage}
\vspace{-4mm}
\end{wrapfigure}
We now employ \algacro{} to \resnetfifty{} on \imagenet{}. Under a similar procedure, \algacro{} first automatically partitions the trainable variables of \resnetfifty{} into PZIGs, and then trains it once by \dhspg{} to automatically construct slimmer models without fine-tuning. We report a performance portfolio under various target group sparsities ranging from 40\% to 70\% and compare with other state-of-the-art methods in Figure~\ref{fig:resnet50_imagenet}. Notably, \algacro{} appears to a Pcareto frontier in terms of balancing top-1 accuracy and FLOPs reduction across different group sparsities. Specifically, at 70\% group sparsity, \algacro's pruned ResNet50 model achieves a significant FLOP reduction 85.5\% while maintaining a competitive top-1 accuracy of 70.3\%, comparable to SFP~\citep{he2018soft} and RBP~\citep{zhou2019accelerate}, but with 3x fewer FLOPs. The variant with 72.3\% top-1 accuracy at 60\% group sparsity rivals CP~\citep{He_2017_ICCV}, DDS-26~\citep{huang2018data}, and RRBP~\citep{zhou2019accelerate} in accuracy, yet is 2-3 times more efficient. 
The slimmer ResNet50 models at 40\% and 50\% group sparsity both achieve an accuracy milestone of around 75\%, outperforming most state-of-the-art models in FLOP reduction. While methods like ResRep~\citep{ding2020lossless}, Group-HS~\citep{yang2019deephoyer}, and GBN-60~\citep{you2019gate} attain over 76\% accuracy, they consume more FLOPs than \algacro{} and lack the automated generality for diverse DNNs.

\paragraph{\bert{} on \squad{}.} In the end, we compare DHSPG versus HSPG on pruning a transformer \bert{}~\citep{NIPS2017_3f5ee243}, evaluated on \squad{}, a question-answering benchmark~\citep{rajpurkar2016squad}. In OTOv2~\citep{chen2023otov2}, the transformer architecture is not supported due to some ONNX compliance issues. Now, \algacro{} could support it due to the recent engineering refactorization of the pruning mode. The results are reported in Table~\ref{table:bert_squad}, showing that DHSPG significantly  outperforms HSPG and ProxSSI~\citep{deleu2021structured} by achieving 83.8\%-87.7\% F1-scores and 74.6\%-80.0\% exact match rates. In constrast, HSPG and ProxSSI reach 82.0\%-84.1\% F1-scores and 71.9\%-75.0\% exact match rates. Such noticeable improvement by DHSPG is driven due to the hybrid training schema in DHSPG to employ distinct updating mechanisms onto separate groups of variables, which avoids trapping around an sub-optimum. In contrast, both ProxSSI and HSPG equip with a unified training schema to penalize the magnitude of all variables, which deteriorates the performance significantly in this experiment. The results well validate the effectiveness of the hybrid training design in DHSPG for typical better generalization. 

\begin{table}[h]
    \centering
    \caption{Structurally pruning Bert on \squad.}
    \label{table:bert_squad}
\begin{minipage}{.7\textwidth}
\resizebox{\textwidth}{!}{
\begin{tabular}{l|c|c|c|c}
	\Xhline{3\arrayrulewidth}
    Method  &\# of Params  & Exact  & F1-score & Inference SpeedUp\\
	\hline
    Baseline &  100\% & 81.0\% & 88.3\% & $1\times$\\
    ProxSSI~\citep{deleu2021structured}  & \hspace{1.3mm}83.4\%$^\dagger$ & 72.3\% & 82.0\% & $1\times$ \\
    HSPG~\citep{chen2021oto} & 91.0\% &  75.0\% & 84.1\% & $1.1\times$ \\
    HSPG~\citep{chen2021oto} & 66.7\% &  71.9\% & 82.0\% & $1.3\times$ \\
    \textbf{DHSPG} (10\% group sparsity) & 93.3\% &  \textbf{80.0\%} & \textbf{87.7\%} & $1.1\times$ \\
    \textbf{DHSPG} (30\% group sparsity) & 80.1\% &  {79.4\%} & {87.3\%} & $1.2\times$ \\
    \textbf{DHSPG} (50\% group sparsity) & 68.3\% &  78.1\% & 86.2\% & $1.3\times$ \\
    \textbf{DHSPG} (70\% group sparsity) & \textbf{55.0\%} &  74.6\% & 83.8\% & $\bm{1.4\times}$ \\
    \hline%
\Xhline{3\arrayrulewidth} 
    \multicolumn{5}{l}{$^\dagger$ Approximate value based on the group sparsity reported in~\citep{deleu2021structured}. }	
\end{tabular}
}
\end{minipage}
    \hspace{1mm}
    \begin{minipage}{.28\textwidth}
    \centering
     \includegraphics[width=\linewidth]{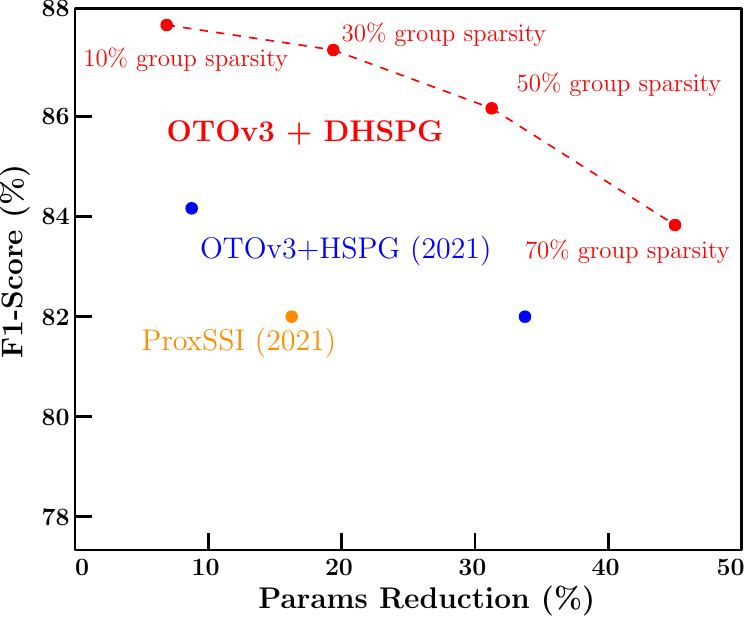}
    \end{minipage}
\end{table}

\subsection{Erasing Operators}

In this section, we employ the erasing mode of \algacro{} to one-shot automatically train and search within general DNNs to construct compact sub-networks with high performance via erasing redundant operators. The numerical demonstrations cover extensive DNNs including \demonet{} shown in Section~\ref{sec.erasing_mode}, \regnet{}~\citep{radosavovic2020designing}, \stackedunets{}~\citep{ronneberger2015u}, \superresnet{}~\citep{he2016deep,ming_zennas_iccv2021}, and \darts{}~\citep{liu2018darts}, over benchmark datasets, including \cifar{}~\citep{Krizhevsky09}, \fashionmnist{}~\citep{xiao2017online}, \imagenet{}~\citep{deng2009imagenet}, \stl{}~\citep{coates2011analysis} and \svnh{}~\citep{netzer2011reading}. 

\begin{table}[t]
    \centering
    \scriptsize
   \caption{Erasing operators on extensive DNNs.}
   \vspace{-3mm}
    \label{table.otov3_various_networks_datasets}
    \vspace{1mm}
	\resizebox{\linewidth}{!}{
	\begin{tabular}{l|c|c|c|c|c}
	\Xhline{3\arrayrulewidth}
    Method  &  Backend  & Dataset & FLOPs (M) & \# of Params (M) & Top-1 Acc. (\%)  \\
    \hline
    Baseline &  \demonet{} & \fashionmnist{} & 209 & 0.82 & 84.9 \\ 
    \textbf{\algacro{} Erasing Mode} & \demonet{} & \fashionmnist{} &  \textbf{107} & \textbf{0.45} & \textbf{84.7}  \\ 
    \hdashline
    Baseline &  StackedUnets & \svnh{} & 184 & 0.80 & 95.3 \\
    \textbf{\algacro{} Erasing Mode} &StackedUnets & \svnh{} & \textbf{115} & \textbf{0.37} & \textbf{96.1} \\ 
    \hdashline
    Baseline & DARTS (8 cells) & STL-10 & 614 & 4.05 & 74.6 \\
    \textbf{\algacro{} Erasing Mode} & DARTS (8 cells) & STL-10 & \textbf{127} & \textbf{0.64} & \textbf{75.1} \\
    \Xhline{3\arrayrulewidth}
	\end{tabular}
	}
 \vspace{-2mm}
\end{table}

\paragraph{\demonet{} on \fashionmnist{}.} We first experiment with the \demonet{} depicted as Figure~\ref{fig:demonet_original} on \fashionmnist{}. \algacro{} automatically establishes the erasing search space of \demonet{} and partitions its trainable variables into EZIGs. \hhspg{} then trains \demonet{} from scratch and computes a solution of high performance and hierarchical group-sparsity, which is further utilized to construct a compact sub-network as presented in Figure~\ref{fig:constructured_sub_network}. As shown in Table~\ref{table.otov3_various_networks_datasets}, compared to the full network, the sub-network utilizes 54\% of parameters and 51\% of FLOPs to achieve a Top-1 validation accuracy 84.7\% which is negligibly lower than the full network by 0.2\%.

\paragraph{\stackedunets{} on \svnh{}.} We then consider a \stackedunets{} over \svnh{}. The \stackedunets{} is constructed by stacking two standard Unets~\citep{ronneberger2015u} with different down-samplers together, as depicted in Figure~\ref{fig:stacked_unet_trace_graph} in Appendix~\ref{appendix.erasing}. We employ \algacro{} to automatically build the erasing search space and train by \hhspg{}. \hhspg{} identifies and projects the redundant EZIGs onto zero and optimizes the remaining important groups to attain excellent performance. As displayed in Figure~\ref{fig:stacked_dependancy_graph}, the right-hand-side Unet is disabled due to \texttt{node-72-node-73-node-74-node-75} identified as redundant.
The path regarding the deepest depth for the left-hand-side Unet, \ie, \texttt{node-13-node-14-node-15-node-19}, is marked as redundant as well. The results by~\algacro{} indicate that the performance gain brought by either composing multiple Unets in parallel or encompassing deeper scaling paths is not significant. \algacro{}'s erasing mode also validates the human design since a single Unet with properly selected depths have achieved remarkable success in numerous applications~\citep{ding2021cdfi,ding2022sparsity,weng2019unet,Geng_2022_CVPR}. Furthermore, as presented in Table~\ref{table.otov3_various_networks_datasets}, the sub-network built by~\algacro{} uses 0.37M parameters and 115M FLOPs which is noticeably lighter than the full \stackedunets{} meanwhile significantly outperforms it by 0.8\% in validation accuracy.

\paragraph{\darts{} (8-Cells) on \stl{}.} We next employ \algacro{} on \darts{} over \stl{}. \darts{} is a complicated network consisting of iteratively stacking multiple cells~\citep{liu2018darts}. Each cell is constructed by spanning a graph wherein every two nodes are connected via multiple operation candidates. \stl{} is an image dataset for the semi-supervising learning, where we conduct the experiments by using its labeled samples. \darts{} has been well explored in the recent years. However, the existing NAS methods studied it based on a \textit{handcrafted} search space beforehand to \textit{locally} pick up one or two important operations to connect every two nodes. We now employ \algacro{}'s erasing mode on an eight-cells \darts{} to \textit{automatically} establish its search space, then utilize \hhspg{} to one shot train it and search important structures \textit{globally} as depicted in Figure~\ref{fig:darts_8cells_dependancy_graph} of Appendix~\ref{appendix.erasing}.  Afterwards, a sub-network is automatically constructed as drawn in Figure~\ref{fig:darts_8cells_subnetwork}. Quantitatively, the sub-network outperforms the full \darts{} in terms of validation accuracy by 0.5\% by using only about 15\%-20\% of the parameters and the FLOPs of the original network (see Table~\ref{table.otov3_various_networks_datasets}).

\begin{table}[h]
\centering
\scriptsize
\caption{Erasing operators over \superresnet{} on \cifar{}.}
\label{table.superresnet_cifar10}
\vspace{-3mm}
\begin{tabular}{lccccc}
\Xhline{3\arrayrulewidth}
\multirow{2}{*}{Method} & \multirow{2}{*}{Type} & \multirow{2}{*}{Search Space} & \multirow{2}{*}{Top-1 Acc (\%)} & \multirow{2}{*}{FLOPs (M)} & \multirow{2}{*}{\# of Params (M)} \\
& & & & \\
\hline
Zen-Score-1M\citep{ming_zennas_iccv2021} & Zero-Shot & ResNet Pool & 96.2 & 159  & {0.99}  \\
Synflow$^\dagger$~\citep{tanaka2020pruning} & Zero-Shot & ResNet Pool  & 95.1 & -- & $\sim$1.0  \\
TE-NAS$^\dagger$~\citep{chen2021neural} & Zero-Shot & ResNet Pool  & 96.1 & --  & $\sim$1.0 \\
NASWOT$^\dagger$~\citep{mellor2021neural} & Zero-Shot & ResNet Pool  & 96.0 & -- & $\sim$1.0 \\
Zen-Score-2M~\citep{ming_zennas_iccv2021} & Zero-Shot & ResNet Pool  & 97.5 & 481 & {1.98} \\
\hdashline
\textbf{\algacro{} Erasing Mode} & Gradient & \superresnet{} & 96.3 & 161 & 1.00  \\
\textbf{\algacro{} Erasing Mode} & Gradient & \superresnet{} & 97.5 & 477 & 1.97 \\
\Xhline{3\arrayrulewidth}
\multicolumn{4}{l}{$^\dagger$ Reported in~\citep{ming_zennas_iccv2021}.}\\ 
\end{tabular}
\vspace{-6mm}
\end{table}

\paragraph{\superresnet{} on \cifar{}.} We switch to a ResNet search space inspired by ZenNAS~\citep{ming_zennas_iccv2021}, referred to as \superresnet{}. ZenNAS~\citep{ming_zennas_iccv2021} uses a ResNet pool to populates massive ResNet candidates and ranks them via zero-shot proxy. We independently construct \superresnet{} by stacking several super-residual blocks with varying depths. Each super-residual blocks contain multiple \texttt{Conv} candidates with kernel sizes as \texttt{3x3}, \texttt{5x5} and \texttt{7x7} separately in parallel. \superresnet{} includes the optimal architecture derived from ZenNAS and aims to discover the most suitable sub-networks using H2SPG over the automated erasing search space. The sub-network produced by~\algacro{} could reach the benchmark over 97\% validation accuracy. 

\begin{table}[ht!]
    \centering
    \caption{\algacro{} over \darts{} on \imagenet{} and comparison with state-of-the-art methods.}
    \vspace{-1mm}
	\label{table.darts_imagenet}
	\resizebox{\linewidth}{!}{
	\begin{tabular}{lccccc}
	\Xhline{3\arrayrulewidth}
	\multirow{2}{*}{Architecture} &  \multicolumn{2}{c}{Test Acc. (\%)} & \multirow{2}{*}{\# of Params (M)} & \multirow{2}{*}{FLOPs (M)}  &  \multirow{2}{*}{Search Method}\\
    \cline{2-3} & Top-1 & Top-5 & & \\
	\hline
	DARTS (2nd order) (\cifar)~\citep{liu2018darts} & 73.3 & 91.3 & 4.7 & 574  & Gradient\\
    SNAS (mild)~\citep{xie2018snas} & 27.3 & 91.2 & 4.3 & 522  & Gradient\\
    P-DARTS (\cifar)~\citep{chen2019progressive} & 75.6 & 92.6 & 4.9 & \cellcolor{yellow!40}{557}   & Gradient\\
    PC-DARTS (\cifar)~\citep{xu2019pc} & 74.9 & 92.2 & 5.3 & 586  & Gradient\\
    ISTA-NAS (\cifar)~\citep{yang2020ista} & 74.9 & 92.3 & 4.8 & 550 &  Gradient\\
    SANAS (\cifar)~\citep{hosseini2022saliency} & 75.2 & 91.7 & -- & --  & Gradient\\
    \hdashline
    ProxylessNAS (\imagenet)~\citep{cai2018proxylessnas} & 75.1 & 92.5 & 7.1 & \cellcolor{red!40}{465}   & Gradient\\
    PC-DARTs (\imagenet)~\citep{xu2019pc} & 75.8 & 92.7 & 5.3 & {597} &  Gradient \\
    ISTA-NAS (\imagenet)~\citep{yang2020ista} & \cellcolor{orange!40}{76.0} & \cellcolor{orange!40}{92.9} & 5.7 & 638  & Gradient \\
    MASNAS (\imagenet)~\citep{lopes2023manas} & 74.7 & -- & \cellcolor{red!40}{2.6} & -- & Multi-Agent \\
    MixPath (\imagenet)~\citep{chu2023mixpath} & \cellcolor{red!40}{77.2} & \cellcolor{red!40}{93.5} & \cellcolor{yellow!40}{5.1} & -- & Gradient\\
    \hdashline
	\textbf{\algacro{}} on \darts{} (\imagenet) & \cellcolor{yellow!40}{75.9} & \cellcolor{yellow!40}{92.8} & \cellcolor{orange!40}{4.9} & \cellcolor{orange!40}{552} & Gradient\\ 
	\Xhline{3\arrayrulewidth}
    \multicolumn{6}{l}{(\cifar{}) / (\imagenet) refer to using either \cifar{} or \imagenet{} for searching architecture. }.
	\end{tabular}
	}
\vspace{-5mm}
\end{table}

\paragraph{\darts{} (14-Cells) on \imagenet{}.} We now present the benchmark \darts{} network stacked by 14 cells on \imagenet{}. \algacro{} automatically figures out the erasing search space, trains by \hhspg{} to figure out redundant erasing minimally removal structures, and constructs a sub-network. (The depictation is ommitted in arxiv version due to exceeding page limit, yet could be found at \href{https://github.com/tianyic/only_train_once}{our github repository}.)  
We observe that the sub-network produced by \algacro{} achieves competitive top-1/5 accuracy compared to other state-of-the-arts as presented in Table~\ref{table.darts_imagenet}. 
Remark that it is {engineeringly} difficult to automatically inject architecture variables and build a multi-level optimization upon the automatic search space. 
Consequently, our accuracy achieved by single-level optimization does not outperform MixPath and ISTA-NAS. We leave further improvement over automated multi-level optimization pipeline as future work. 

\paragraph{Ablation Study (\regnet{} on \cifar).} We finally conduct ablation studies over \regnet{}~\citep{radosavovic2020designing} on \cifar{} to demonstrate the necessity and efficacy of hierarchical sparse optimizer \hhspg{} compared to the existing non-hierarchical sparse optimizers, which is the key to the erasing mode. Without loss of generality, we employ \algacro{} over the  RegNet-800M  which has accuracy 95.01\% on \cifar{}, and compare with the latest variant of HSPG, \ie, \dhspg{}~\citep{chen2023otov2}. 
We evaluate them with varying target hierarchical group sparsity levels in problem~(\ref{prob.main.erasing}) across a range of $\{0.1, 0.3, 0.5, 0.7, 0.9\}$. As other experiments, \algacro{} automatically constructs its search space, trains via \hhspg{} or \dhspg{}, and establishes the sub-networks without fine-tuning.  The results are from three independent tests under different random seeds, and reported in Table~\ref{table.otov3_regnet_cifar10}.
\begin{table}[h]
    \centering
    \scriptsize
   \caption{\algacro{} on \regnet{} on \cifar{}.}
   \vspace{-2mm}
    \label{table.otov3_regnet_cifar10}
\begin{minipage}{.68\textwidth}
    \resizebox{\linewidth}{!}{
	\begin{tabular}{ cccccc}
	\Xhline{3\arrayrulewidth}
    \multirow{2}{*}{Backend}   & \multirow{2}{*}{Method} & \multirow{2}{*}{Optimizer} & Target  & \multirow{2}{*}{\# of Params (M)} & \multirow{2}{*}{Top-1 Acc. (\%)}  \\
    & & & Sparsity  & &  \\
    \hline
      &  &  & 0.1 & $5.56 \pm\  0.02$  & $95.26 \pm\ 0.13$ \\
     &  &  & 0.3 & (3.40, \xmark, \xmark)  & (95.01, \xmark, \xmark) \\ 
    RegNet-800M  & \textbf{\algacro{}} & \dhspg{} & 0.5 & (\xmark, \xmark, \xmark) & (\xmark, \xmark, \xmark) \\ 
      &  &  & 0.7 & (\xmark, \xmark, \xmark) & (\xmark, \xmark, \xmark) \\ 
      &  &  & 0.9 & (\xmark, \xmark, \xmark) & (\xmark, \xmark, \xmark) \\
    \hdashline
     & & & 0.1 & $5.58 \pm\  0.01$  & $95.30 \pm\ 0.10$ \\
      & &  & 0.3 & $3.54 \pm\  0.15$ & $95.08 \pm\ 0.14$ \\ 
    RegNet-800M  & \textbf{\algacro{}} & \textbf{\hhspg{}} & 0.5 & $1.83 \pm\ 0.09$ & $94.61 \pm\ 0.19$ \\ 
      &  &  & 0.7 & $1.16 \pm\ 0.12$ & $91.92 \pm\ 0.24$ \\ 
     &  &  & 0.9 & $0.82 \pm\ 0.17$ & $87.91 \pm\ 0.32$ \\
    \Xhline{3\arrayrulewidth}
	\end{tabular}
	}
\end{minipage}
\hspace{2mm}
\begin{minipage}{.29\textwidth}
    \centering
    \includegraphics[width=\linewidth]{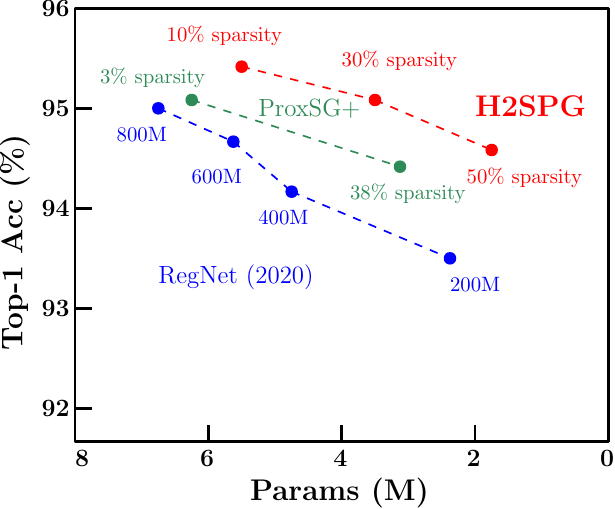}
\end{minipage}
\end{table}

\noindent
\contour{black}{Sub-networks by~\algacro{} versus Full Networks.}  The sub-networks under varying hierarchical group sparsity levels computed by~\algacro{} with \hhspg{} exhibits the Pareto frontier comparing with the benchmark \regnet{}s. Notably, the sub-networks under sparsity levels of 0.1 and 0.3 outperform the full RegNet-800M. Furthermore, the ones with 0.5 sparsity level outperforms the RegNet(200M-600M), despite utilizes significantly fewer parameters while achieves higher accuracy.

\noindent
\contour{black}{\hhspg{} versus Other Sparse Optimizers.} \dhspg{} often fails when confronts with reasonably large target sparsity levels, denoted by the symbol \xmark. The underlying reason lies in its design, which solely treats problem~(\ref{prob.main.erasing}) as an independent and disjoint structured sparsity problem. By disregarding the hierarchy within the network, \dhspg{} easily generates invalid sub-networks. Conversely, \hhspg{} takes into account the network hierarchy and successfully addresses problem~(\ref{prob.main.erasing}). We also compare with a proximal method equipping with our hierarchical search phase, \ie, ProxSG+. Its performance is not competitive to \hhspg{} due to their ineffective sparse exploration ability~\citep{dai2023adaptive}.
\section{Conclusions}\label{sec.conclusion}

We propose \algacro{}, a pioneering framework, that automatically trains a general DNN only once and compress it into a more compact counterpart by either structurally pruning or erasing operators. For different compression mode, \algacro{} automatically generates the corresponding search spaces along with distinct and novel dependency graph analysis and zero-invariant group partitions. Later on, two novel sparse optimizers \dhspg{} or \hhspg{} are employed to jointly identify redundant structures and train the important ones to high-performance. Remarkably, \hhspg{} also stands as the first optimizer to address the hierarchical structured sparsity problems for deep learning applications. \algacro{} significantly reduces the human efforts upon the existing structured pruning and NAS works, opens new directions, and establishes benchmarks regarding the automated general DNN compression.


\acks{We thank Microsoft LLC for supporting the series of Only-Train-Once (OTO) framework. Zhihui Zhu acknowledges support from NSF grant CCF-2240708.
}

\newpage
\bibliography{main}

\newpage
\appendix
\section{Visualization for Pruning Dependency Graph}\label{appendix.erasing}


\begin{figure}[H]
\centering
\includegraphics[height=0.9\textheight]{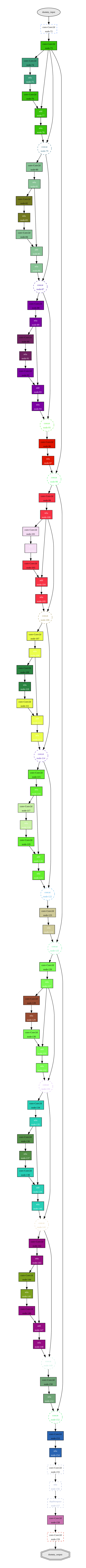}
\caption{Pruning dependency graph for CARNx2.}
\label{fig:carnx2_pruning_dep}
\end{figure}



\begin{figure}[H]
\centering
\includegraphics[height=0.9\textheight]{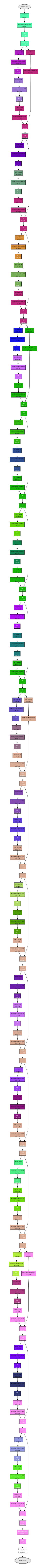}
\caption{Pruning dependency graph for ResNet50.}
\label{fig:resnet_pruning_dep}
\end{figure}

\begin{figure}[H]
\centering
\includegraphics[height=0.9\textheight]{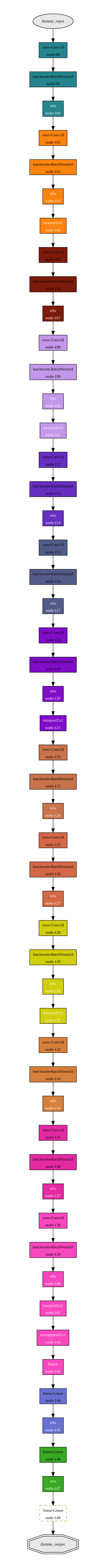}
\caption{Pruning dependency graph for VGG16-BN.}
\label{fig:vgg16_pruning_dep}
\end{figure}

\newpage
\section{Visualization for Erasing Mode}\label{appendix.erasing}
In this appendix, we present visualizations generated by the erasing mode of \algacro{} library to provide more intuitive illustrations of the architectures tested in the paper. The visualizations include trace graphs, erasing dependency graphs, identified redundant erasing minimally removal structures, and constructed sub-networks. To ensure clear visibility, we highly recommend \textbf{zooming in with an upscale ratio of 500\%} to observe finer details and gain a better understanding of the system.

\begin{figure}[H]
    \centering
    \begin{subfigure}[b!]{\textwidth}
    \centering
    \includegraphics[height=0.7\textheight]{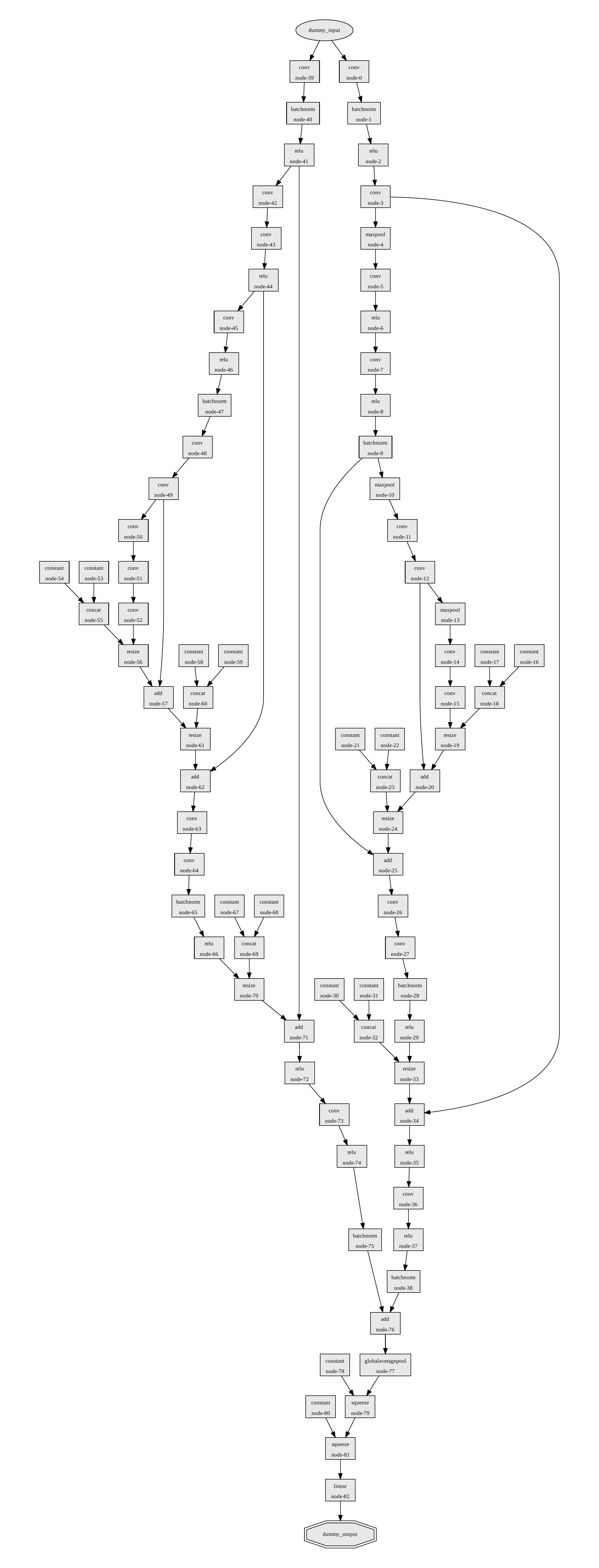}
    \caption{\stackedunets{} trace graph.}
    \label{fig:stacked_unet_trace_graph}
    \end{subfigure}
    \caption{\stackedunets{} illustrations drawn by~\algacro{}.}
    \label{fig:my_label}
\end{figure}
\begin{figure}[H]
\ContinuedFloat
    \begin{subfigure}[b!]{\textwidth}
    \centering
    \includegraphics[height=0.9\textheight]{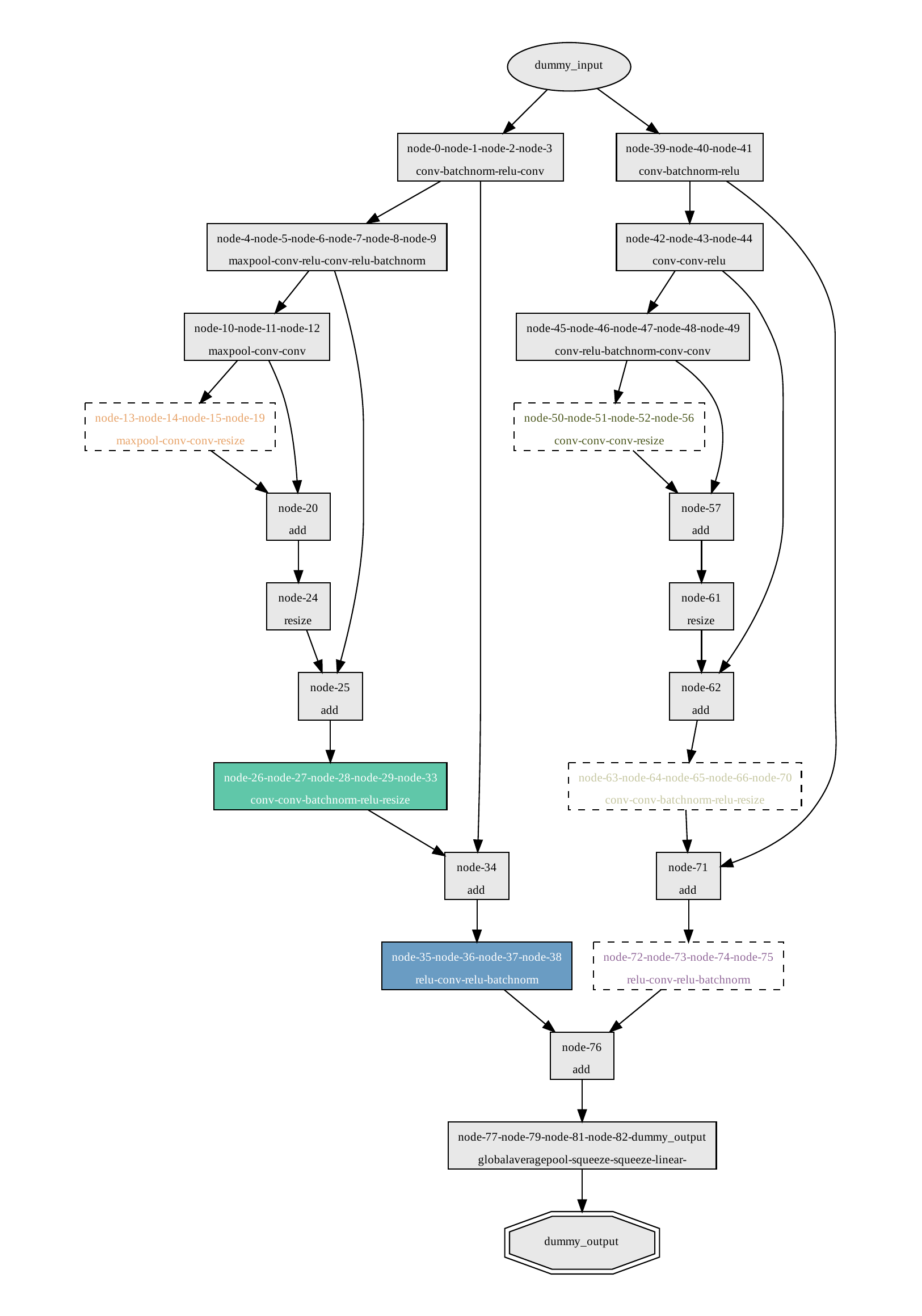}
    \caption{\stackedunets{} erasing dependency graph with identified redundant removal structures.}
    \label{fig:stacked_dependancy_graph}
    \end{subfigure}
    \caption{\stackedunets{} illustrations drawn by~\algacro{}.}
\end{figure}
\begin{figure}[H]
\ContinuedFloat
    \begin{subfigure}[b!]{\textwidth}
    \centering
    \includegraphics[height=0.9\textheight]{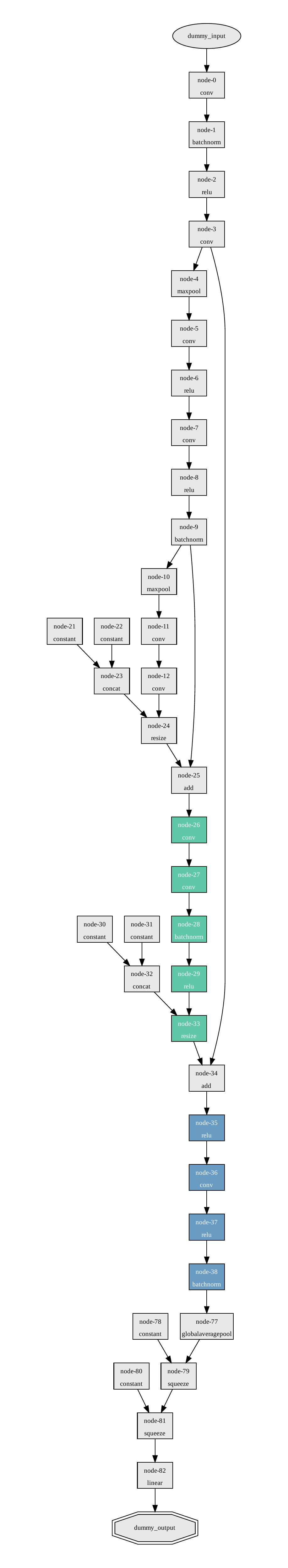}
    \caption{Constructed sub-network upon \stackedunets{}.}
    \label{fig:stacked_subnetwork}
    \end{subfigure}
    \caption{\stackedunets{} illustrations drawn by~\algacro{}.}
\end{figure}

\begin{figure}[H]
    \centering
    \begin{subfigure}[b!]{\textwidth}
    \centering
    \vspace{1mm}
    \includegraphics[height=0.95\textheight]{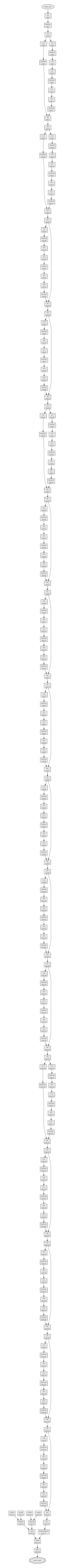}
    \caption{\regnet{} trace graph.}
    \label{fig:regnet_trace_graph}
    \end{subfigure}
    \caption{\regnet{} illustrations drawn by~\algacro{}.}
    \label{fig:my_label}
\end{figure}
\begin{figure}[H]
\ContinuedFloat
    \begin{subfigure}[b!]{\textwidth}
    \centering
    \includegraphics[height=0.95\textheight]{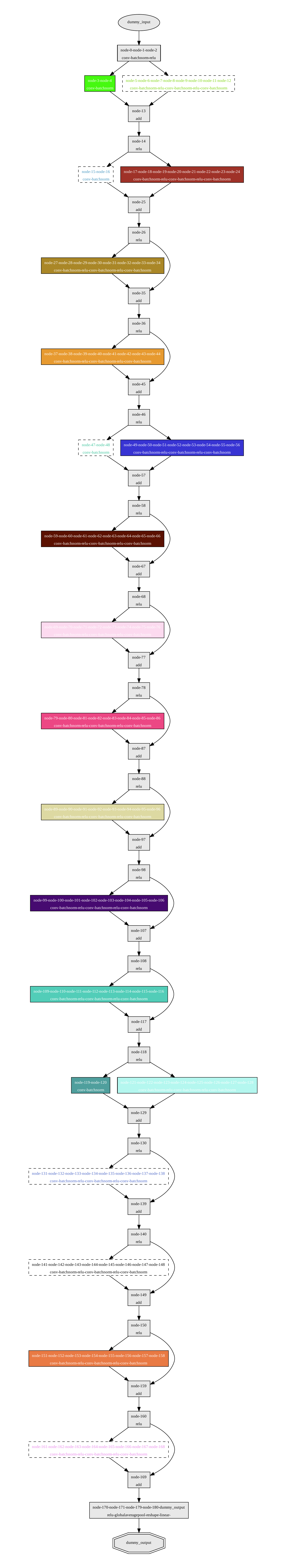}
    \caption{\regnet{} erasing dependency graph with identified redundant removal structures.}
    \label{fig:regnet_redundant_structures}
    \end{subfigure}
    \caption{\regnet{} illustrations drawn by~\algacro{}.}
\end{figure}
\begin{figure}[H]
\ContinuedFloat
    \begin{subfigure}[b!]{\textwidth}
    \centering
    \includegraphics[height=0.95\textheight]{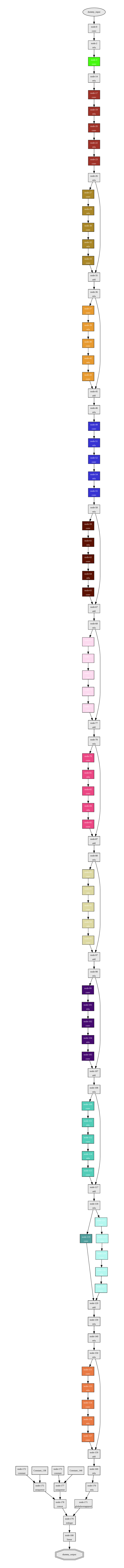}
    \caption{Constructed sub-network upon \regnet{}.}
    \label{fig:regnet_sub_network}
    \end{subfigure}
    \caption{\regnet{} illustrations drawn by~\algacro{}.}
    \label{fig:regnet}
\end{figure}

\end{document}